\documentclass{article}

\usepackage{microtype}
\usepackage{graphicx}
\usepackage{booktabs} 

\usepackage{hyperref}

\usepackage[accepted]{icml2021}

\usepackage[page]{appendix}

\usepackage{natbib}
\usepackage{amsmath}
\usepackage{amssymb}
\usepackage{mathtools}
\usepackage{caption}
\usepackage{subcaption}
\usepackage[dvipsnames]{xcolor}
\usepackage{ulem}
\usepackage{multirow}
\usepackage{arydshln}
\usepackage{tikz}
\usepackage[geometry]{ifsym}
\usepackage{comment}
\graphicspath{ {./figures/} }

\DeclareMathOperator*{\argmax}{arg\,max}
\DeclareMathOperator*{\argmin}{arg\,min}
\DeclareMathOperator*{\Exp}{\mathbb{E}}

\definecolor{hotpink}{rgb}{0.858, 0.188, 0.478}

\icmltitlerunning{GMAC: A Distributional Perspective on Actor-Critic Framework}

\begin{document}

\twocolumn[
\icmltitle{GMAC: A Distributional Perspective on Actor-Critic Framework}



\icmlsetsymbol{equal}{*}

\begin{icmlauthorlist}
\icmlauthor{Daniel Wontae Nam}{ml2}
\icmlauthor{Younghoon Kim}{ml2}
\icmlauthor{Chan Y. Park}{ml2}
\end{icmlauthorlist}

\icmlaffiliation{ml2}{KC Machine Learning Lab, Seoul, Korea}

\icmlcorrespondingauthor{Daniel}{dwtnam@kc-ml2.com}

\icmlkeywords{Machine Learning, Reinforcement Learning, ICML 2021}

\vskip 0.3in
]



\printAffiliationsAndNotice{}  

\begin{abstract}
In this paper, we devise a distributional framework on actor-critic as a solution to distributional instability, action type restriction, and conflation between samples and statistics.
We propose a new method that minimizes the Cram\'er distance with the multi-step Bellman target distribution generated from a novel Sample-Replacement algorithm denoted SR($\lambda$), which learns the correct value distribution under multiple Bellman operations.
Parameterizing a value distribution with Gaussian Mixture Model further improves the efficiency and the performance of the method, which we name GMAC.
We empirically show that GMAC captures the correct representation of value distributions and improves the performance of a conventional actor-critic method with low computational cost, in both discrete and continuous action spaces using Arcade Learning Environment (ALE) and PyBullet environment.
\end{abstract}

\section{Introduction}

The ability to learn complex representations via neural networks has enjoyed success in various applications of reinforcement learning (RL), such as pixel-based video gameplays \citep{mnih15}, the game of Go \citep{silver16}, robotics \citep{levine16}, and high dimensional controls like humanoid robots \citep{lillicrap16, schulman15}.
Starting from the seminal work of Deep Q-Network (DQN) \citep{mnih15}, the advance in value prediction network, in particular, has been one of the main driving forces for the breakthrough.

Among the milestones of the advances in value function approximation, distributional reinforcement learning (DRL) further develops the scalar value function to a distributional representation.
The distributional perspective offers various benefits by providing more information on the characteristics and the behavior of the value.
One such benefit is the preservation of multimodality in value distributions, which leads to more stable learning of the value function \citep{bellemare17}.

\begin{figure}[tb]
\centering
\begin{subfigure}{\columnwidth}
\centering
\includegraphics[width=0.5\columnwidth]{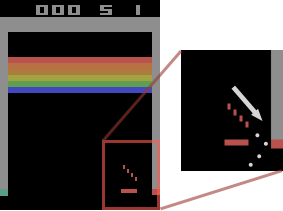}
\caption{The observation input}
\end{subfigure}
\begin{subfigure}{\columnwidth}
\centering
\includegraphics[width=\columnwidth]{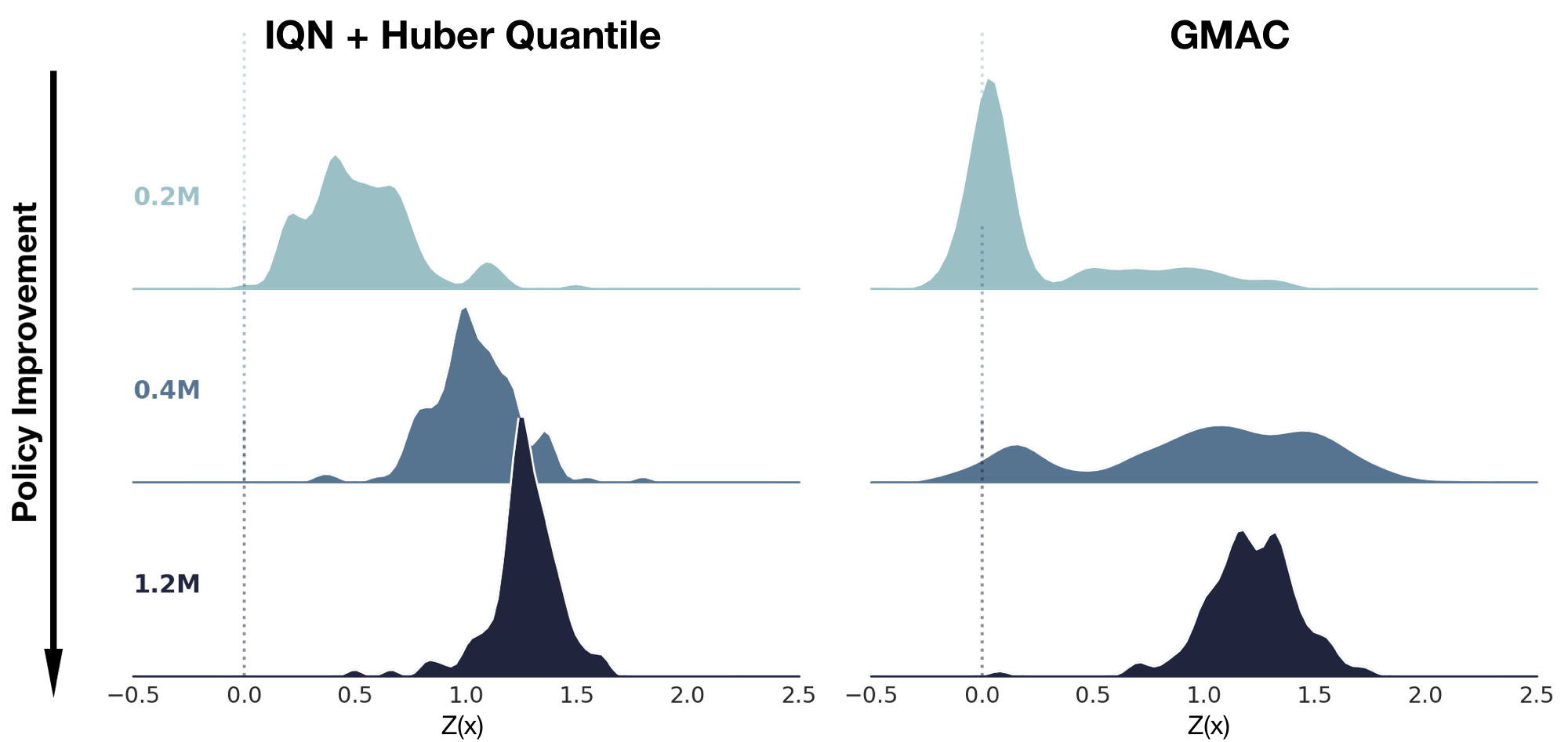}
\caption{The evaluated value distributions}
\end{subfigure}
\vskip 0.1in
\caption{
Modality of value distribution during the learning process of Breakout-v4.
(a) An arrow is added in the inset to indicate the ball's direction of travel.
The episode reaches a terminal state if the paddle misses the ball.
(b) Learned probability density functions of the value distributions of IQN + Huber-quantile (left) vs. GMAC (Gaussian mixture model + energy distance) (right) for a same policy when trained on \{0.2, 0.4, 1.2\}M frames.
As the policy improves, the probability of losing a turn ($V=0$) should decrease while the probability of earning scores ($V > 0$) increases.
Note that the modality transition from $V=0$ is clearly captured by GMAC.
}
\vskip -0.1in
\label{fig:modality}
\end{figure}

Despite the development, several issues remain, hindering DRL from becoming a robust framework.
First, a theoretical instability exists in the control setting of value-based DRL methods \citep{bellemare17}.
Second, previous DRL algorithms are limited to a single type of action space, either discrete \citep{bellemare17, dabney17, dabney18} or continuous \citep{barth-maron18, singh20}.
Third, a common choice of loss function for DRL is the Huber quantile regression loss, which is vulnerable to conflation between samples and statistics without an imputation strategy \citep{rowland19}.

While the instability and action space issue can be avoided simply by applying a general actor-critic framework \citep{williams88, williams92, sutton99}, practical methods critical to actor-critic framework such as TD($\lambda$) have not been established in the distributional perspective.
Therefore we suggest a novel sample-replacement algorithm denoted by SR($\lambda$) to generate multi-step Bellman target distribution with high efficiency.
Furthermore, we avoid the conflation problem by directly learning samples through minimizing the Cram\'er distance between distributions.

As proven in \citep{rowland19}, using an imputation strategy can help DRL methods to learn a more accurate representation of a distribution.
However, many actor-critic methods are designed to use multi-step returns such as the $\lambda$-return \citep{watkins89} for which the imputation strategy can be a computational burden.
Therefore, we instead construct the multi-step returns from samples and parameters to avoid the necessity of imputation.
We propose to parameterize the value distribution as a Gaussian mixture model (GMM), and minimize the Cram\'er distance between the distributions.
When combining GMM with the energy distance, a specific case of the Cram\'er distance, we can derive an analytic solution and obtain unbiased sample gradients at a much lower computational cost compared to the method using the Huber quantile loss.
We call our framework GMAC (Gaussian mixture actor-critic).

We present experimental results to demonstrate how GMAC can successfully solve the three problems of DRL.
Firstly, we illustrate that GMAC is a competitive actor-critic framework by showing that the framework outperforms its baseline algorithms in the Atari games\citep{ale}.
Secondly, the experiments on the continuous control tasks in PyBullet environments \citep{pybullet} show that the same framework can be used for both tasks with discrete and continuous action spaces.
Lastly, we share the FLOP measurement results to show that the accurate representation of value distributions can be learned with less computational cost.

\section{Related Works}

\citet{bellemare17} has shown that the distributional Bellman operator derived from the distributional Bellman equation is a contraction in a maximal form of the Wasserstein distance.
Based on this point, \citet{bellemare17} proposed a categorical distributional model, C51, which is later discussed to be minimizing the Cram\'er distance in the projected distributional space \citep{rowland18, bellemare19, qu19}.
\citet{dabney17} proposed quantile regression-based models, QR-DQN, which parameterizes the distribution with a uniform mixture of Diracs and uses sample-based Huber quantile loss \citep{huber1964}.
\citet{dabney18} later expanded it further so that a full continuous quantile function can be learned through the implicit quantile network (IQN).
\citet{yang19} then further improved the approximation of the distribution by adjusting the set of quantiles.
\citet{choi19} suggested parameterizing the value distribution using Gaussian mixture and minimizing the Tsallis-Jenson divergence as the loss function on a value-based method.
Outside of RL, \citet{bellemare17cramer} proposed to use Cram\'er distance in place of Wasserstein distance used in WGAN due to its unbiasedness in sample gradients \citep{arjovsky17}.

There have been many applications of the distributional perspective, which exploit the additional information from value distribution.
\citet{dearden98} modeled parametric uncertainty and \citet{morimura10a, morimura10b} designed a risk-sensitive algorithm using a distributional perspective, which can be seen as the earliest concept of distributional RL. 
\citet{mavrin19a} utilized the idea of the uncertainty captured from the variance of value distribution.
\citet{nikolov18} has also utilized the distributional representation of the value function by using information-directed-sampling for better exploration of the value-based method.
While multi-step Bellman target was considered \citep{Hessel2018RainbowCI}, the sample-efficiency was directly addressed by combining multi-step off-policy algorithms like Retrace($\lambda$) \citep{gruslys17}.

Just as C51 has been expanded deep RL to distributional perspective, \citet{barth-maron18} studied a distributional perspective on DDPG \citep{lillicrap16}, an actor-critic method, by parameterizing a distributional critic as categorical distribution and Gaussian mixture model.
\citet{singh20} has further expanded the work by using an implicit quantile network for the critic. 
Several works \citep{duan20, kuznetsov20, ma20} have proposed a distributional version of the soft-actor-critic (SAC) framework to address the error from over-estimating the value.
These works mainly focused on combining a successful distributional method with a specific RL algorithm.
To this end, this paper aims to suggest a more general method that can extend any actor-critic to the distributional perspective.

\section{Distributional Reinforcement Learning}
\label{sec:DistRL}
We consider a conventional RL setting, where an agent's interaction with its environment is described by a Markov Decision Process (MDP) ($\mathcal{X}, \mathcal{A}, R, P, \gamma$), where $\mathcal{X}$ and $\mathcal{A}$ are state and action spaces, $R(x,a)$ is the stochastic reward function for a pair of state $x$ and action $a$, $P(x'|x,a)$ is the transition probability of observing $x'$ given the pair $(x,a)$, and $\gamma \in$ (0,1) is a time discount factor. 
A policy $\pi(\cdot|x)$ maps a state $x \in \mathcal{X}$ to a probability distribution over actions $a \in \mathcal{A}$.

The objective of RL is to maximize the expected return, $\mathbb{E}[G_t]$ where $G_t=\sum_{t=0}^{\infty} \gamma^t R(x_t, a_t)$ is the sum of discounted rewards from state $x_t$ given a policy $\pi$ at time $t$.
Then for any state $x_t$, the value $V$ and state-action value $Q$ under the given policy $\pi$ can be defined as
\begin{align}
V(x_t) &= \mathbb{E}[G_t \mid X=x_t], \\
\quad Q(x_t, a_t) &= \mathbb{E}[G_t \mid X=x_t, A=a_t].
\end{align}
A recursive relationship in the value in terms of the reward and the 
transition probability is described by the Bellman equation \citep{bellman58} given by
\begin{align}
Q(x, a) = \mathbb{E}[R(x, a)] + \gamma \mathbb{E}_{a' \sim \pi, x' \sim P}\left[Q(x', a') \right],
\end{align}
where the first expectation is calculated over a given state-action pair $(x, a)$ and the second expectation is taken over the next possible states $x' \sim P(\cdot | x, a)$ and actions $a' \sim \pi(\cdot | x)$.

DRL extends the Bellman equation to an analogous recursive equation, termed the distributional Bellman equation \citep{morimura10a, morimura10b, bellemare17}, using a distribution of the possible sum of discounted rewards $Z(x, a)$:
\begin{align}
\label{eq:DBE}
Z^\pi(x, a) &\overset{D}{=} R(x, a) + \gamma Z^\pi(X', A'),
\end{align}
where $\overset{D}=$ denotes having equal distributions and $Q(x, a) = \mathbb{E}[Z(x, a)]$.
Then $Z$ is learned through distributional Bellman operator $\mathcal{T}^\pi$ defined as
\begin{align}
\label{eq:dist_bellman_op}
\mathcal{T}^{\pi} Z(x, a) :\overset{D}{=} R(x, a) + \gamma P^\pi Z(x, a)
\end{align}
where $P^{\pi}: \mathcal{Z} \rightarrow \mathcal{Z}$ is a state transition operator under policy $\pi$, $P^\pi Z(x, a) \overset{D}{:=} Z(X', A')$, where $X'\sim P(\cdot | x, a)$ and $A' \sim \pi(\cdot | X')$.
Analogously, the distributional Bellman optimality operator $\mathcal{T}$ can be defined as
\begin{align}
\label{eq:DBO}
\mathcal{T} Z(x,a) \overset{D}{:=} R(x,a) + \gamma Z(X', \argmax_{a'} \mathbb{E} [Z(X',a')]).
\end{align}
The distributional Bellman operator has been proven to be a $\gamma$-contraction in a maximal form of Wasserstein distance \citep{bellemare17}, which has a practical definition given by 
\begin{align}
d_p(U, V) = \left( \int_0^1 \lvert F^{-1}_U(\omega) - F^{-1}_V (\omega) \rvert^p d\omega \right)^{1/p},
\end{align}
where $U, V$ are random variables and $F_U, F_V$ are their cumulative distribution functions (cdf).

However, unlike the distributional Bellman operator, the distributional Bellman optimality operator is not a contraction in any metric \citep{bellemare17}, causing an instability where the distance $d_p(\mathcal{T}Z_1, \mathcal{T}Z_2)$ between some random variables $Z_1, Z_2$ may not converge to a unique solution.
This issue has been discussed in \citet{bellemare17}, with an example of oscillating value distribution caused by a specific tie-breaker design of the \textit{argmax} operator.

The instability can be removed simply by learning the value distribution under the evaluation setting of the Bellman operation described in \eqref{eq:dist_bellman_op}.
This, on the other hand, poses a new problem on the RHS of \eqref{eq:dist_bellman_op}: the next state-action value distribution $Z(X', A')$ becomes a mixture distribution of all possible state-action value distributions, the computation of which can be infeasible for value-based methods in continuous action space.
We avoid this issue by directly approximating the state value distribution $Z(X')$ instead of the state-action value distribution $Z(X', A')$.
This lets us to use the general actor-critic policy gradient,
\begin{align}
\nabla J(\theta) = \mathbb{E}\left[ A_t \nabla \ln(\pi_\theta(x_t,a_t)) \right],
\end{align}
where the advantage $A_t$ may be estimated using the temporal-difference (TD) error between expectations of the value distributions, or using generalized advantage estimation \citep{schulman16} in a similar manner.
At this point, we are left with estimating  multi-step distributional Bellman target while avoiding the data conflation problem \citep{rowland19}.

\section{Algorithm}
\label{sec:Algorithm}

\begin{figure*}[tb]
\centering
\resizebox{2.0\columnwidth}{!}{
\begin{tikzpicture}
\node [inner sep=0pt] (figure) at (5,0) {\includegraphics[width=2.0\columnwidth]{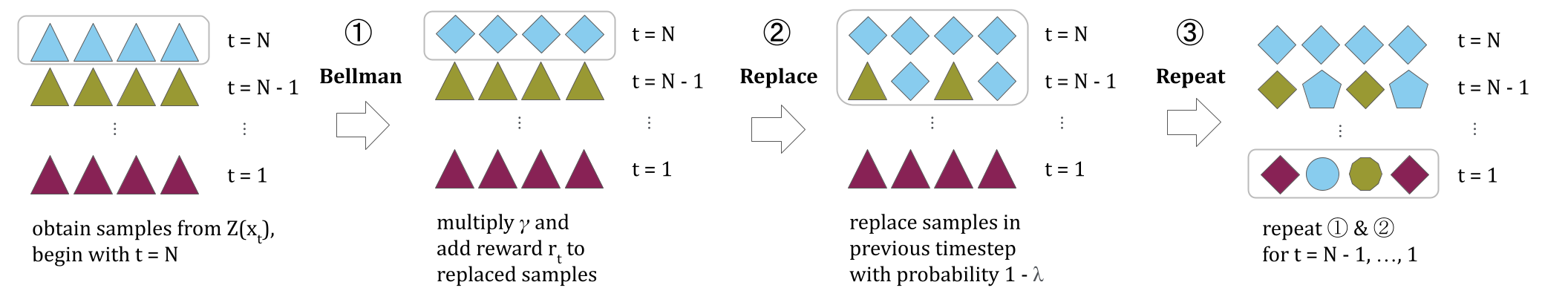}};
\node [align=left] (input) at (-1.7,2) {\textbf{Input:} \\ $\{Z(x_1), \dots, Z(x_N)\}$};
\node [align=left] (output) at (11,2) {\textbf{Output:} \\ $\{Z_0^{(\lambda)}, \dots, Z_{N-1}^{(\lambda)}\}$};
\end{tikzpicture}
}
\caption{
A visual representation of the SR($\lambda$) algorithm for generating $N$ distributional Bellman targets from given a trajectory of length $N$.
Set of dirac samples ($m=4$) are represented by the colored shapes.
Shapes of same colors denote that they are transformed from the same $Z(x_t)$ and different shapes denote numbers of Bellman operation applied to a specific sample.
Each row of samples in the rightmost figure represent the samples the $\lambda$-return Bellman target $Z_{t-1}^{(\lambda)}$ for each time step.
}
\label{fig:sr-lambda-diagram}
\end{figure*}

\subsection{SR(\texorpdfstring{$\lambda$}{lambda}): Sample-Replacement for \texorpdfstring{$\lambda$}{lambda}-return Distribution}
\label{subsec:sample-replacement}

Here we introduce SR($\lambda$), a novel method for estimating the multi-step distributional Bellman target, analogous to the TD($\lambda$) in the case of scalar value functions.
The actor-critic method is a temporal-difference (TD) learning method in which the value function, the critic, is learned through the TD error defined by the difference between the TD target given by $n$-step return, $G_t^{(n)} = \sum_{i=1}^n \gamma^{i-1} r_{t+i} + \gamma^n V(x_{t+n})$, and the current value estimate $V(x_t)$.
A special case of TD method, called TD($\lambda$) \citep{sutton88}, generates a weighted average of $n$-step returns for the TD target, also known as the $\lambda$-return,
\begin{align}
\label{eq:g_lambda}
\begin{split}
G_t^{(\lambda)} = (1 - \lambda) \sum_{n=1}^\infty \lambda^{n-1} G_t^{(n)}, \quad \lambda \in [0,1],
\end{split}
\end{align}
to mitigate the variance and bias trade-off between Monte Carlo and the TD(0) return to enhance data efficiency.

An important piece of SR$(\lambda)$ is the use of a random variable $Z_t^{(\lambda)}$ that is the distributional analogue of the $\lambda$-return, which we propose in the following.
First let us define a random variable $\tilde{G}$ whose sample space is the set of all $n$-step returns, $\{G_t^{(1)}, \ldots, G_t^{(\infty)}\}$ with the probability distribution given by
\begin{align}
\label{eq:p_g_lambda}
\Pr[\tilde{G} = G_t^{(n)}] = (1-\lambda) \lambda^{n-1}.
\end{align}
Then, \eqref{eq:g_lambda} is same as the expectation of the random variable $\tilde{G}$.
Similar to $G_t^{(n)}$, we define $n$-step approximation of the value distribution as
\begin{align}
\begin{split}
Z_t^{(n)} :\overset{D}{=} \sum_{i=0}^{n-1} \gamma^{i} R(x_{t+i}, a_{t+i}) + \gamma^n Z(x_{t+n}),
\end{split}
\end{align}
where $\mathbb{E} [ Z_t^{(n)} ] = G_t^{(n)}$.
Then we can imagine a random variable $\tilde{Z}$ whose sample space is a set of all $n$-step approximations, $\{Z_t^{(1)}, \ldots, Z_t^{(\infty)}\}$ , which are random variables as well.
However, unlike $\tilde{G}$ whose expectation is a scalar value, i.e. the weighted mean of its supports, the expectation of $\tilde{Z}$ is a random variable that has a distribution equal to the mixture of distributions of $Z_t^{(n)}$.
To avoid the ambiguity of ``an expectation of a random variable of random variables", we define the distributional analogue of \eqref{eq:p_g_lambda} in terms of cdfs:
\begin{align}
\label{eq:p_Fn}
\Pr [\tilde{F} = F_{Z_t^{(n)}}] = (1-\lambda) \lambda^{n-1}.
\end{align}
$F_{Z_t^{(n)}}$ denotes the cdf of the $n$-step return $Z_t^{(n)}$, and $\tilde{F}$ is a random variable over the set of $F_{Z_t^{(n)}} \in \{F_{Z_t^{(0)}}, \dots , F_{Z_t^{(\infty)}}\}$.
Then using \eqref{eq:p_Fn}, we can successfully define the expectation of $\tilde{F}$ as a linear combination of $F_{Z_t^{(n)}}$
\begin{align}
\label{eq:exp_F}
\mathbb{E} [\tilde{F}] = (1 - \lambda) \sum_{n=1}^\infty \lambda^{n-1} F_{Z_t^{(n)}}.
\end{align}
Let us define a random variable $Z_t^{(\lambda)}$ that has $\Exp[\tilde{F}]$ as its cdf, i.e. the probability distribution of $Z_t^{(\lambda)}$ is a mixture distribution of the probability distributions of $Z_t^{(n)}$'s.
Then the expectation of $Z_t^{(\lambda)}$ and the expectation of $Z_t^{(n)}$ have an analogous relationship to \eqref{eq:g_lambda} (see \autoref{sec:E_z_t_lambda}), meaning that the expectation of $Z_t^{(\lambda)}$ is equal to the $\lambda$-return.

Note that, in practice, collecting infinite horizon trajectory is infeasible and thus the truncated sum is often used \citep{cichosz95, seijen11}:
\begin{align}
F_{Z_t^{(\lambda)}} = (1 - \lambda) \sum_{n=1}^N \lambda^{n-1} F_{Z_t^{(n)}} + \lambda^{N} F_{Z_t^{(N)}}.
\end{align}
Given a trajectory of length $N$, naively speaking, finding $Z_t^{(\lambda)}$ for each time step requires finding $N$ different $Z_t^{(n)}$. As a result, we need to find total of $O(N^2)$ different distributions to find $Z_t^{(\lambda)}$ for all states in the given trajectory. But the number of distributions to find reduces to $O(N)$ when we create approximations of $Z_t^{(n)}$ beforehand and reuse them for calculating $Z_t^{(\lambda)}$ for each time step. 

One choice among such approximations is to use a mixture of diracs from the sample values, as described in \citep{dabney17}:
\begin{align}
\label{eq:empirical_distribution}
Z_t^{(n)} \approx Z_\theta(x_t) := \frac{1}{m} \sum_{i=1}^m \delta_{\theta_i(x_t)},
\end{align}
where $\theta: \mathcal{X} \rightarrow \mathbb{R}^m$ is some parametric model.
In this case, we can approximate the distribution of $Z_t^{(\lambda)}$ by aggregating the samples from each $Z_\theta(x_t)$ with probability $\lambda^{n-1}$.
Since the total set of samples remains unchanged during the calculation, we can create the $N$ different $Z_t^{(\lambda)}$ in a single sweep by replacing a portion of samples for each time step, which leads to the name of our method Sample-Replacement, or SR($\lambda$).

Figure \ref{fig:sr-lambda-diagram} describes SR($\lambda$) schematically.
The approximated distribution of the $\lambda$-returns, $Z_t^{(\lambda)}$, for the last state in a trajectory is simply given by the samples of the last value distribution $Z_\theta(x_t)$.
Then traversing the trajectory in a reversed time order, we replace each of the samples with a new sample from the earlier time step with a probability of $1 - \lambda$.
The replaced collection of samples is used as the approximation for $Z_t^{(\lambda)}$ in that time step.
We iterate this process until the beginning of the trajectory to create a total of $N$ approximations.
A more detailed description of the algorithm can be found in Algorithm \autoref{alg:sr_lambda}.
We further propose to apply SR($\lambda$) to the parameters of GMM instead of dirac samples in the following sections.
There exists a closed-form solution for minimizing the Cram\'er distance between Gaussian mixtures, which enables us to create unbiased gradients at a lower computational cost compared to when using samples or statistics.
In the case of statistics, \citet{rowland19} has shown that one should use an imputation strategy on the statistics to acquire samples of the distribution, which may add significant computational overhead.
\begin{algorithm}[tb]
\caption{SR($\lambda$)}
\label{alg:sr_lambda}
\begin{algorithmic}
\STATE {\bfseries Input:} Trajectory of states and value distributions $\{(x_1, Z_1), \ldots, (x_N, Z_N)\}$ of length $N$, discount factor $\gamma$, weight parameter $\lambda$
\STATE {\bfseries Output:} Set of $\lambda$-returns $\{Z_0^{(\lambda)}, \ldots, Z_{N-1}^{(\lambda)}\}$
\STATE $\boldsymbol{X} \leftarrow$ Collect $m$ samples $\{X_1, \ldots, X_m\}$ from $Z_{N}$ \;
\FOR{$t=N-1$ {\bfseries to} $0$}
\STATE  $\boldsymbol{X} \leftarrow r_t + \gamma \boldsymbol{X}$ \hfill\COMMENT{Bellman operation}
\STATE $Z_t^{(\lambda)} \leftarrow \sum_{i=1}^m \delta_{X_i}$ 
\STATE \hfill\COMMENT{empirical distribution using $m$ Diracs}
\STATE $\boldsymbol{X}' \leftarrow$ Collect $m$ samples $\{X'_1, \ldots, X'_m\}$ from $Z_t^{(\lambda)}$ \;
\FOR{$i = 1$ {\bfseries to} $m$}
\STATE$X_i \leftarrow X'_i$ with probability $1 - \lambda$
\ENDFOR
\ENDFOR
\end{algorithmic}
\end{algorithm}

\subsection{Cram\'er Distance}

Let $P$ and $Q$ be probability distributions over $\mathbb{R}$.
If we define the cdf of $P, Q$ as $F_P,  F_Q$ respectively, the $l_p$ family of divergence between $P$ and $Q$ is
\begin{align}
l_p(P,Q) := \left( \int_{-\infty}^{\infty} |F_P(x) - F_Q(x)|^p dx \right)^{1/p}.
\end{align}
When $p=2$, it is termed the Cram\'er distance.
The distributional Bellman operator in the evaluation setting is a $|\gamma|^{1/p}$-contraction mapping in the Cram\'er metric space \citep{rowland19, qu19}, whose worked out proof can also be found in Appendix \ref{sec:contraction-proof}.

A notable characteristic of the Cram\'er distance is the unbiasedness of the sample gradient,
\begin{align}
\Exp_{X \sim Q} \nabla_\theta l_2^2(\hat{P}_m, Q_\theta) = \nabla_\theta l_2^2(P,Q_\theta)
\end{align}
where $\hat{P}_m := \frac{1}{m} \sum_{i=1}^{m} \delta_{X_i}$ is the empirical distribution, and $Q_\theta$ is a parametric approximation of a distribution.
The unbiased sample gradient makes it suitable to use Cram\'er distance with stochastic gradient descent method and empirical distributions for updating the value distribution.

\citet{szekely} showed that, in the univariate case, the squared Cram\'er distance is equivalent to one half of \textit{energy distance} ($l_2^2(P,Q) = \frac{1}{2} \mathcal{E}(P,Q)$) defined as 
\begin{align}
\begin{split}
\mathcal{E}(P,Q) &:= \mathcal{E}(U,V) \\
&= 2 \Exp \|U-V\|_2 - \Exp \|U-U'\|_2 - \Exp \|V-V'\|_2,
\end{split}
\end{align}
where $U, U'$ and $V, V'$ are random variables that follow $P, Q$, respectively.
Then, energy distance can be approximated using the random samples of $U$ and $V$.

\subsection{Energy Distance between Gaussian Mixture Models}
\label{subsec:gmm}
We take a step further to enhance the approximation accuracy and computational efficiency by considering the parameterized model of the value distribution as a GMM \citep{choi19, barth-maron18}.
Following the same assumption used for \eqref{eq:empirical_distribution}, the approximation using GMM is given using parametric models $\mu, \sigma, w: \mathcal{X} \rightarrow \mathbb{R}^K$ 
\begin{align}
\begin{split}
Z_\theta(x_t) &\sim \sum^{K}_{i=1} w_i(x_t)\, \mathcal{N}(z \vert \mu_i(x_t), \sigma_i(x_t)^2), \\
&\text{where} \: \sum_{i=1}^K w_i(x_t) = 1.
\end{split}
\end{align}

If random variables $U, V$ follow the distributions $P,Q$ parameterized as GMMs, the energy distance has the following closed-form
\begin{align}
\label{eq:energy_distance}
\begin{split}
\mathcal{E}(U,V) &= 2 \delta(U,V) - \delta(U,U') - \delta(V,V'), \\
\text{where} \: \\
\delta(U,V) &= \sum_{i,j} w_{ui} w_{vj} \Exp \left[ \lvert Z_{ij} \rvert \right], \\
Z_{ij} & \sim \mathcal{N}(z \vert \mu_{ui}-\mu_{vj}, \sigma_{ui}^2 + \sigma_{vj}^2).
\end{split}
\end{align}
Here, $\mu_{xi}$ refers to the $i^{th}$ component for random variable $X$ and same applies for $\sigma$ and $w$ for both $U$ and $V$.
The closed-form solution of the energy distance defined in \eqref{eq:energy_distance} has a computational advantage over sample-based approximations like the Huber quantile loss. When using the GMM, the analytic approximation of \eqref{eq:empirical_distribution} can be derived as
\begin{align}
\begin{split}
Z_t^{(\lambda)} &\sim \sum_{n=1}^\infty (1-\lambda) \lambda^{n-1} \sum_{k=1}^K w_{nk}\, \mathcal{N}(z \vert \mu_{nk}, \sigma_{nk}^2)\\
&\approx \frac{1}{m} \sum_{i=1}^m \mathcal{N}(z \vert \mu_{nk}, \sigma_{nk}^2), \\
n & \sim \text{Geo}(1-\lambda) \\
k \, \vert \, n & \sim \text{Categorical}(w_{n1}, \dots ,w_{nk})
\end{split}
\end{align}
where $\mu_{nk}$ refers to $k^{th}$ component of $\mu_{Z_t^{(n)}}$ for simplicity of notation.
This is equivalent to having a mixture of $m$ Gaussians, thus we can simply perform sample replacement on the parameters $(\mu, \sigma^2)$, instead of realizations of the random variables as in \eqref{eq:empirical_distribution}.
Then, the distance function described in \eqref{eq:energy_distance} can easily be applied.

When bringing all the components together, we have a distributional actor-critic framework with SR($\lambda$) that minimizes the energy distance between Gaussian mixture value distributions.
Comprehensively, we call this method GMAC.
A brief sketch of the algorithm is shown in \autoref{sec:pseudocode-gmac}.

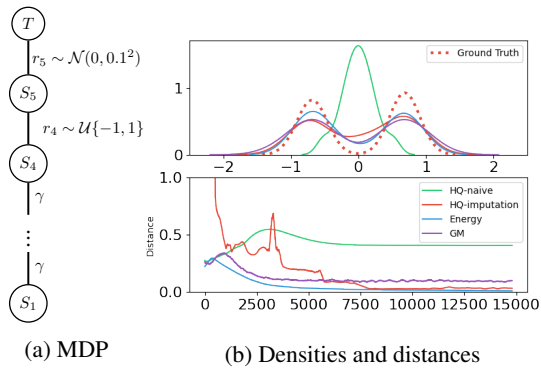
\begin{figure}[t]
\centering
\begin{subfigure}{0.2\columnwidth}
\centering
\resizebox{1.2\columnwidth}{!}{
\begin{tikzpicture}[scale=0.5, line width=2pt,
thin, node distance=1.0cm, auto,
blank/.style = {draw=none, rectangle},
node/.style = {draw, thick, circle, minimum size=7mm},
edge/.style = {draw=arrow, thick, black}]

\node[node] (s1) at (0,0){$S_1$};
\node[blank] (dots) at (0,3){\Large $\vdots$};
\node[node] (s4) at (0,6){$S_4$};
\node[node] (s5) at (0,9){$S_5$};
\node[node] (term) at (0,12){$T$};
\node[blank] (g1) at (0.5,1.5){$\gamma$};
\node[blank] (g2) at (0.5,4.5){$\gamma$};
\node[blank] (r1) at (2.8,7.5){$r_4 \sim \mathcal{U}\{-1, 1\}$};
\node[blank] (r2) at (2.5,10.5){$r_5 \sim \mathcal{N}(0,0.1^2)$};

\path [very thick]
(s1) edge [right] node {} (dots)
(dots) edge [right] node {} (s4)
(s4) edge [above] node {} (s5)
(s5) edge [above] node {} (term);
\end{tikzpicture}
}
\caption{MDP}
\end{subfigure}
\begin{subfigure}{.7\columnwidth}
\centering
\includegraphics[width=\columnwidth]{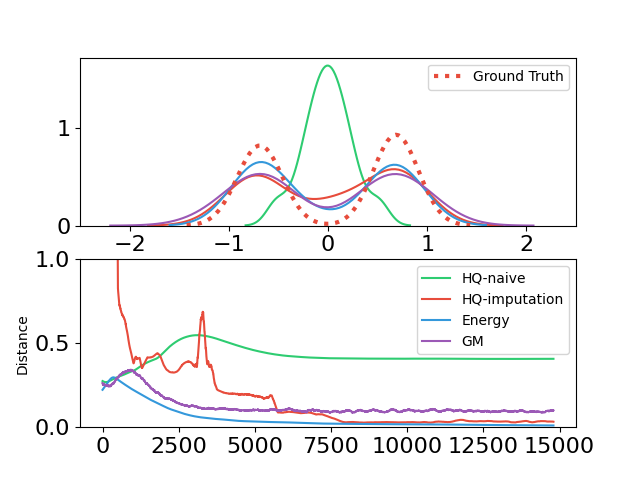}
\caption{Densities and distances}
\end{subfigure}
\vskip 0.1in
\caption{
(a) An environment with five states and stochastic rewards with expected value of zero.
(b) The probability density functions (above) and the energy distance between the ground truth and the estimated distributions (below) for tabular methods with Huber-quantile, Huber-quantile with imputation, energy distance, and GMM with energy distance.
}
\vskip -0.1in
\label{fig:modality-toy}
\end{figure}

\section{Experiments}
\label{sec:Experiments}
In this section, we present experimental results for three different distributional versions of Proximal Policy Optimization (PPO) with SR($\lambda$): IQAC (IQN + Huber quantile), IQAC-E (IQN + energy distance), and GMAC (GMM + energy distance), in the order of the progression of our suggested approach.
The performance of the scalar version of PPO with value clipping \citep{schulman16} is used as the baseline for comparison. 
Details about the loss function of each method can be found in \autoref{sec:loss_function}.
For a fair comparison, we keep all common hyperparameters consistent across the algorithms except for the value heads and their respective hyperparameters (see \autoref{sec:implementation-details}).

The results demonstrate three contributions of our proposed DRL framework:
1) the ability to correctly capture the multimodality of value distributions,
2) generalization to both discrete and continuous action spaces, and
3) significantly reduced computational cost.

\paragraph{Representing Multimodality}
As discussed throughout \autoref{sec:Algorithm}, we expect minimizing the Cram\'er distance to produce a correct depiction of a distribution without using an imputation strategy.
First, we demonstrate this with a simple value regression problem for an MDP of five sequential states, as shown in \autoref{fig:modality-toy} (a).
The reward function $r_i$ of last two state $S_i$ is stochastic, with $r_4$ from a uniform discrete distribution and $r_5$ from a normal distribution.
Then the value distribution of $S_1$ should be bimodal with expectation of zero (\autoref{fig:modality-toy} (b)).
In this example, minimizing the Huber-quantile loss (labeled as HQ-naive) of dirac mixture underestimates the variance of $S_1$ due to conflation and does not capture the locations of the modes.
By applying an imputation strategy as suggested in \citet{rowland19}, a slight improvement on the underestimation of variance can be seen.
On the other hand, both dirac mixture and GMM, labeled as Energy and GM respectively in the figure, show that minimizing the energy distance converges to correct mode locations.
For a fair comparison, GMM uses one-third of the parameters used in the dirac mixtures as its number of mixtures.
More details about the experimental setup and further results can be found in \autoref{sec:extra_experiments}.
The comparison is extended to complex tasks such as the Atari games, of which an example result is shown in \autoref{fig:modality}, and additional visualizations of the value distribution during the learning process from different games can be found in \autoref{sec:extra_experiments}.


\begin{figure*}[ht]
\centering
\begin{subfigure}{\columnwidth}
\centering
\includegraphics[width=.71\columnwidth]{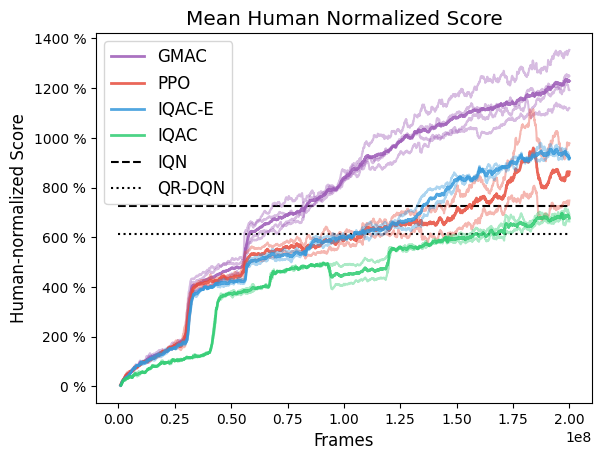}
\caption{ALE mean human-normlized score}
\end{subfigure}
\begin{subfigure}{\columnwidth}
\centering
\includegraphics[width=.71\columnwidth]{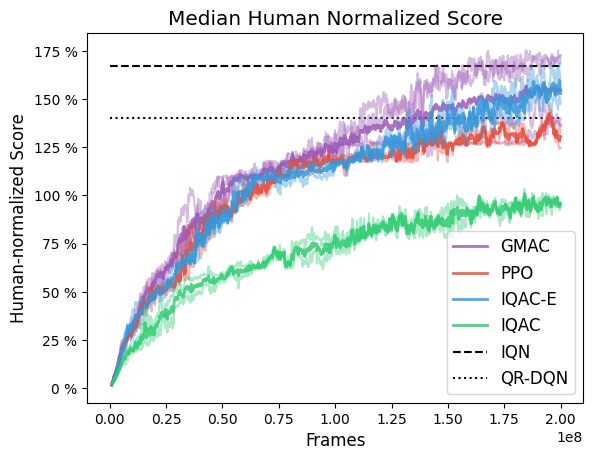}
\caption{ALE median human-normlized score}
\end{subfigure}
\begin{subfigure}{.6\columnwidth}
\centering
\includegraphics[width=\columnwidth]{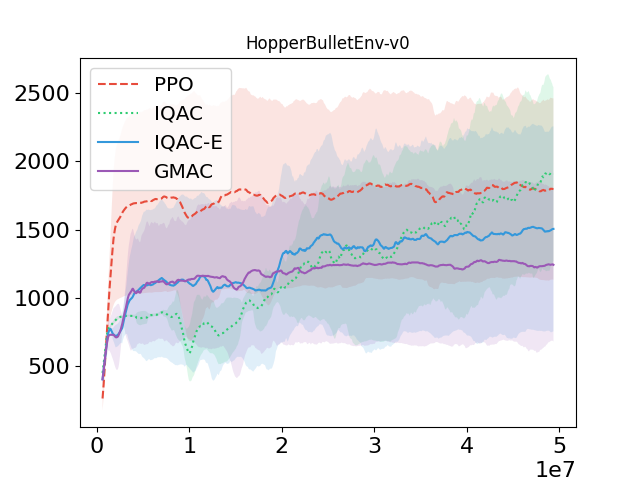}
\caption{Hopper}
\end{subfigure}
\begin{subfigure}{.6\columnwidth}
\centering
\includegraphics[width=\columnwidth]{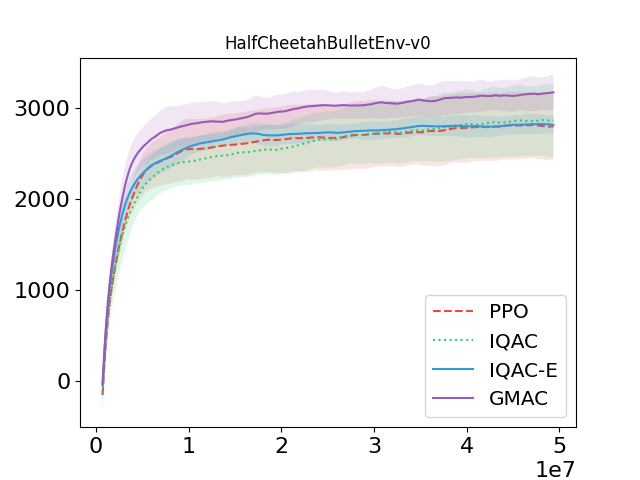}
\caption{HalfCheetah}
\end{subfigure}
\begin{subfigure}{.6\columnwidth}
\centering
\includegraphics[width=\columnwidth]{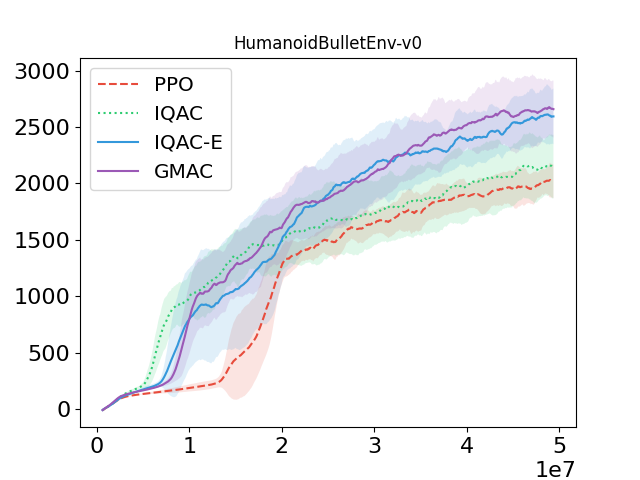}
\caption{Humanoid}
\end{subfigure}
\vskip 0.1in
\caption{
Learning curves for Atari games from ALE and 3 continuous control tasks from PyBullet during training.
The scores for IQN and QR-DQN are taken from \citet{dabney18}.
For Atari, GMAC was run for 3 seeds, PPO and other baselines were run for 2 seeds.
Results of selected Atari games run for 5 seeds can be found in \autoref{sec:extra_experiments}.
For PyBullet, all algorithms were run for 5 seeds. 
}
\vskip -0.1in
\label{fig: atari-pybullet-experiment}
\end{figure*}
\paragraph{Discrete and Continuous Action Spaces}
The human normalized score for 57 Atari games in ALE \citep{ale} is presented in \autoref{fig: atari-pybullet-experiment}.
The results show that GMAC outperforms its scalar baseline PPO and other known distributional methods IQN and QR-DQN in mean scores.
On the other hand, in the median scores, GMAC places between IQN and QR-DQN.
The results tell us that GMAC significantly outperforms the value-based distributional methods in some of the Atari games while its overall performance is competitive.
Another clear distinction is that there is a significant decrease in performance when implicit quantile network is used with Huber-quantile loss for the critic with same architecture with same hyperparameters. In contrast, using energy distance as the loss function ensures non-degenerative performance.
The learning curves for each of 61 Atari games, including ones that did not have human scores, can be found in \autoref{sec:extra_experiments}.

The same exact algorithm is taken to continuous control task of PyBullet environments \citep{pybullet}, with the changes only made in the hyperparameters and policy parameters, from softmax logits to mean and variance of normal distribution.
Without any continuous-control specific modifications made, our methods produce competitive performance compared to the scalar version PPO with slight improvements in the hard tasks such as HumanoidBulletEnv-v0.
More results can be found in \autoref{sec:extra_experiments}.

\paragraph{Computational Cost}
\autoref{table:flops} shows the number of parameters and the number of floating-point operations (FLOPs) required for a single inference and update step of each agent.
We emphasize three points here.
Firstly, the implicit quantile network requires more parameters due to the intermediate embeddings of random quantiles.
Secondly, the difference between the FLOPs for a single update in IQAC and IQAC-E indicates that the proposed energy distance requires less computation than the Huber quantile regression.
Lastly, the results for GMAC show that using GMM can greatly reduce the cost even to match the numbers of PPO while having improved performance.

\begin{table}[ht]
\caption{FLOP measurement results for a single process in Breakout-v4}
\label{table:flops}
\centering
\begin{tabular}{cccc}
\toprule
\multicolumn{1}{c}{{\multirow{2.5}{*}{\bf Algorithm}}} &
\multicolumn{1}{c}{{\multirow{2.5}{*}{\bf Params (M)}}} &
\multicolumn{2}{c}{\bf FLOPs (G)} \\
\cmidrule(lr){3-4}
\multicolumn{1}{c}{} &
\multicolumn{1}{c}{} &
\multicolumn{1}{c}{\bf Inference} &
\multicolumn{1}{c}{\bf Update} \\
\midrule

PPO   & 0.44        & 1.73               & 5.19             \\ \hdashline \rule{0pt}{10pt}
IQAC  & 0.52        & 2.98               & 12.98            \\
IQAC-E & 0.52        & 2.98               & 8.98             \\
GMAC  & \textbf{0.44} & \textbf{1.73}  & \textbf{5.27}      \\

\bottomrule
\end{tabular}
\end{table}

\begin{figure}[t]
\centering
\begin{subfigure}{0.7\columnwidth}
\centering
\includegraphics[width=\columnwidth]{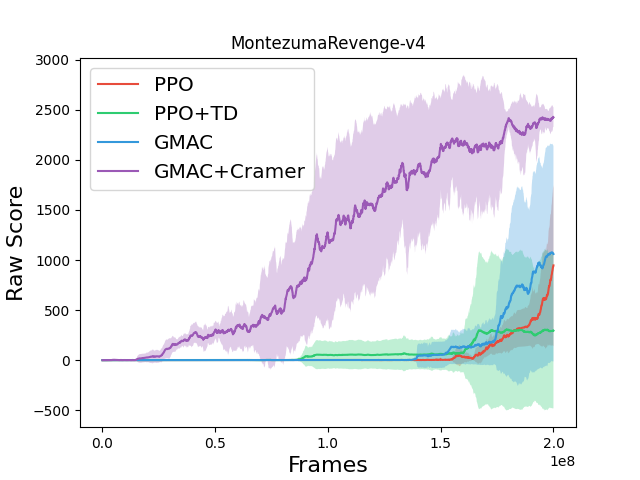}
\caption{Montezuma's revenge with intrinsic rewards}
\end{subfigure}
\vskip 0.1in
\caption{
Learning curve of Montezuma's Revenge using modality information as intrinsic reward.
For a fair comparison, both TD error and energy distance used for intrinsic rewards are normalized in a similar manner to that of RND \citep{burda19} to keep the scales at a similar level.
}
\vskip -0.1in
\label{fig:exploration}
\end{figure}

\paragraph{Using Distributions}
By capturing the correct modes of a value distribution, an additional degree of freedom on top of the expected value can be accurately obtained, from which richer information can be derived to distinguish states by their value distributions.
In particular, the extra information may be utilized as an intrinsic motivation in sparse-reward exploration tasks.
To demonstrate the plausibility of such application, we compare using Cram\'er distance between value distributions as intrinsic reward to using TD error between scalar value estimates in a sparse reward environment of Montezuma's Revenge in \autoref{fig:exploration}, which shows a clear improvement in performance.

\section{Conclusion}

In this paper, we have developed the distributional perspective of the actor-critic framework which integrates the SR($\lambda$) method, Cram\'er distance, and Gaussian mixture models for improved performance in both discrete and continuous action spaces at a lower computational cost.
Furthermore, we show that our proposed method can capture the correct modality in the value distribution, while the extension of the conventional method with the stochastic policy fails to do so.

Capturing the correct modality of value distributions can improve the performance of various policy-based RL applications that exploit statistics from the value distribution.
Such applications may include training risk-sensitive policies and learning control tasks with sparse rewards that require heavy exploration, where transient information from the value distribution can give benefit to the learning process.
We leave further development of these ideas as future works.

\bibliography{main}
\bibliographystyle{icml2021}





\newpage
\onecolumn
\begin{appendices}

\section{Discussion on the choice of Proximal Policy Optimization as a baseline}
\label{sec:conservative-policy-iteration-to-ppo}

A general learning process of RL can be described using policy iteration, which consists of two iterative phases: policy evaluation and policy improvement \citep{sutton98}.
In policy iteration, the value function is assumed to be exact, meaning that given policy, the value function is learned until convergence for the entire state space, which results in a strong bound on the rate of convergence to the optimal value and policy \citep{puterman94}.

But the exact value method is often infeasible from resource limitation since it requires multiple sweeps over the entire state space.
Therefore, in practice, the value function is approximated, i.e.\ it is not trained until convergence nor across the entire state space on each iteration.
The approximate version of the exact value function method, also known as \textit{asynchronous value iteration}, still converges to the unique optimal solution of the Bellman optimality operator. 
However, the Bellman optimality only describes the limit convergence, and thus the best we can practically consider is to measure the improvement on each update step. 

\citet{bertsekas96} have shown that, when we approximate the value function $V_\pi$ of some policy $\pi$ with $\Tilde{V}$, the lower bound of a greedy policy $\pi'$ is given by
\begin{align}
V_{\pi'}(x) \ge V_{\pi}(x) - \frac{2\gamma \varepsilon}{1-\gamma},
\end{align}
where $\varepsilon=\text{max}_x |\Tilde{V}(x) - V_{\pi}(x)|$ is the $L_{\infty}$ error of value approximation $\Tilde{V}$.
This means a greedy policy from an approximate value function guarantees that its exact value function will not degrade more than $\frac{2\gamma \varepsilon}{1-\gamma}$.
However, there is no guarantee on the improvement, i.e.\ $V_{\pi'}(x) > V_{\pi}(x)$ \citep{kakade02}.

As a solution to this issue, \citet{kakade02} have proposed a policy updating scheme named \textit{conservative policy iteration}, 
\begin{align}
\pi_{new}(a|x) = (1-\alpha) \pi_{old}(a|x) + \alpha \pi'(a|x),
\end{align}
which has an explicit lower bound on the improvement
\begin{align}
\label{eq:lower_bound}
\eta(\pi_{new}) &\ge L_{\pi_{old}}(\pi_{new}) - \frac{2 \epsilon \gamma}{(1-\gamma)^2} \alpha^2,
\end{align}
where $\epsilon = \max_x \left\lvert \mathbb{E}_{\pi'} \left[ A_\pi(x,a) \right] \right\rvert$, $A_\pi(x, a) = Q(x, a) - V(x)$ is the advantage function, $\eta(\pi)$ denotes the expected sum of reward under the policy $\pi$, 
\begin{align}
\eta(\pi) = \mathbb{E}\left[ \sum_{t=0}^{\infty} \gamma^{t}R(x_t, a_t) \right],
\end{align}
and $L_{\pi_{old}}$ is the local approximation of $\eta$ with the state visitation frequency under the old policy.

From the definition of distributional Bellman optimality operator in \eqref{eq:DBO}, one can see that the lower bound in \eqref{eq:lower_bound} also holds when $\pi'$ is greedy with respect to the expectation of the value distribution, i.e., $\mathbb{E}_{x' \sim P} [Z(x',a')]$. Thus the improvement of the distributional Bellman update is guaranteed in expectation under conservative policy iteration, and the value functions are guaranteed to converge in distribution to a fixed point by $\gamma$-contraction.

\citet{schulman15} takes this further, suggesting an algorithm called trust region policy optimization (TRPO), which extends conservative policy iteration to a general stochastic policy by replacing $\alpha$ with Kullback-Leibler (KL) divergence between two policies,
\begin{align}
D_{KL}^{max}(\pi, \Tilde{\pi}) = \max_x D_{KL}\left(\pi(\cdot|x) \Vert \Tilde{\pi}(\cdot|x)\right).
\end{align}
Then, the newly formed objective is to maximize the following, which is a form of constraint optimization with penalty:
\begin{align}
\hat{\Exp}_t \left[ \frac{\pi(a_t|x_t)}{\tilde{\pi}(a_t|x_t)} \hat{A}_t - \beta D_{KL}(\pi(\cdot|x_t), \tilde{\pi}(\cdot|x_t)) \right]
= \hat{\Exp}_t \left[ r_t(\pi)\hat{A}_t - \beta D_{KL}(\pi(\cdot|x_t), \tilde{\pi}(\cdot|x_t)) \right].
\end{align}
where $r(\pi)$ refers to the ratio $r(\pi) = \frac{\pi(a_t|x_t)}{\tilde{\pi}(a_t|x_t)}$.
However, in practice, choosing a fixed penalty coefficient $\beta$ is difficult and thus \citet{schulman15} uses hard constraint instead of the penalty.

\begin{align}
& \max_{\theta} \hat{\Exp}_t \left[  r_t(\pi)\hat{A}_t\right] \\
& \quad \text{s.t.} \ D_{KL}(\pi(\cdot|x_t), \tilde{\pi}(\cdot|x_t)) \leq \delta
\end{align}

\citet{schulman17} simplifies the loss function even further in proximal policy optimization (PPO) by replacing KL divergence with ratio clipping between the old and the new policy with the following:
\begin{align}
L^{CLIP} = \hat{\Exp}_t \left[ \min \left( r_t(\pi)\hat{A}_t, \text{clip}(r_t (\pi), 1 - \epsilon, 1 + \epsilon)\hat{A}_t \right) \right].
\end{align}
Thus, by using PPO as the baseline, we aim to optimize the value function via unique point convergence of distributional Bellman operator for a policy being approximately updated under the principle of conservative policy.

\section{Expectation value of \texorpdfstring{$Z_t^{(\lambda)}$}{distributional lambda-return}}
\label{sec:E_z_t_lambda}
Continuing from \eqref{eq:exp_F}, let us define a random variable that has a cumulative distribution function of $\Exp[\tilde{F}_Z]$ as $Z_t^{(\lambda)}$.
Then, its cumulative distribution function is given by 
\begin{align}
F_{Z_t^{(\lambda)}} = (1 - \lambda) \sum_{n=1}^\infty \lambda^{n-1} F_{Z_t^{(n)}}.
\end{align}
If we assume that the support of $Z_t^{(\lambda)}$ is defined in the extended real line $[-\infty, \infty]$,
\begin{align}
\label{eq:lambda_pdf}
\mathbb{E}[Z_t^{(\lambda)}] &= \int^\infty_0 \left(1 - F_{Z_t^{(\lambda)}}\right) dz - \int_{-\infty}^0 F_{Z_t^{(\lambda)}} dz \\
&= \int^\infty_0 \left(1 - (1 - \lambda) \sum_{n=1}^\infty \lambda^{n-1} F_{Z_t^{(n)}}\right) dz - \int_{-\infty}^0 (1 - \lambda) \sum_{n=1}^\infty \lambda^{n-1} F_{Z_t^{(n)}} dz \\
&= (1 - \lambda) \sum_{n=1}^\infty \lambda^{n-1} \left[ \int_0^\infty \left(1 - F_{Z_t^{(n)}}\right) dz - \int_{-\infty}^0 F_{Z_t^{(n)}} dz \right] \\
&= (1 - \lambda) \sum_{n=1}^\infty \lambda^{n-1} G_t^{(n)} = G_t^{(\lambda)}.
\end{align}
Thus we can arrive at the desired expression of $\mathbb{E}[Z_t^{(\lambda)}] = G_t^{(\lambda)}$.



\section{Distributional Bellman operator as a contraction in Cram\'er metric space}\label{sec:contraction-proof}

The Cram\'er distance possesses the following characteristics (detailed derivation of each can be found in \citep{bellemare17cramer}):
\begin{align}
\label{eq:cramer_character}
&l_p(A+X, A+Y) \le l_p(X,Y),
&l_p(cX, cY) \le |c|^{1/p}l_p(X,Y).
\end{align}
Using the above characteristics, the Bellman operator in $l_p$ divergence is
\begin{align}
\begin{split}
l_p\left(\mathcal{T}^\pi Z_1(x,a), \mathcal{T}^\pi Z_2(x,a)\right)
&= l_p(R(x,a) + \gamma P^\pi Z_1(x,a), R(x,a) + \gamma P^\pi Z_2(x,a)) \\
&\le |\gamma|^{1/p} l_p(P^\pi Z_1(x,a), P^\pi Z_2(x,a)) \\
&\le |\gamma|^{1/p} \sup_{x',a'} l_p(Z_1(x',a'), Z_2(x',a')).
\end{split}
\end{align}
Substituting the result into the definition of the maximal form of the Cram\'er distance yields
\begin{align}
\begin{split}
\Bar{l}_p(\mathcal{T}^\pi Z_1, \mathcal{T}^\pi Z_2) &= \sup_{x,a} l_p(\mathcal{T}^\pi Z_1(x,a), \mathcal{T}^\pi Z_2(x,a)) \\
&\le |\gamma|^{1/p} \sup_{x',a'} l_p(Z_1(x',a'), Z_2(x',a')) \\
&= |\gamma|^{1/p} \Bar{l}_p (Z_1, Z_2).
\end{split}
\end{align}
Thus the distributional Bellman operator is a $|\gamma|^{1/p}$-contraction mapping in the Cram\'er metric space, which was also proven in \citet{rowland19}.

Similar characteristics as in \eqref{eq:cramer_character} can be derived for the energy distance
\begin{align}
&\mathcal{E}(A+X, A+Y) \le \mathcal{E}(X,Y),
&\mathcal{E}(cX, cY) = c \mathcal{E}(X,Y),
\end{align}
showing that the distributional Bellman operator is a $\gamma$-contracton in energy distance
\begin{align}
\mathcal{E}(\mathcal{T}^\pi Z_1, \mathcal{T}^\pi Z_2) \le \gamma \mathcal{E}(Z_1, Z_2).
\end{align}

\section{Loss Functions}
\label{sec:loss_function}
As in other policy gradient methods, our value distribution approximator models the distribution of the value, $V(x_t)$, not the state-action value $Q(x_t, a_t)$, and denote it as $Z_\theta(x_t)$ parametrized with $\theta$, whose cumulative distribution function is defined as
\begin{align}
F_{Z_\theta(x_t)} = \sum_{a \in \mathcal{A}} \pi(a, x_t) F_{Z(x_t, a)}.
\end{align}
Below, we provide the complete loss function of value distribution approximation for each of the cases used in experiments (\autoref{sec:Experiments}).

\subsection{Implicit Quantile + Huber quantile (IQAC)}
For the value loss of IQAC, we follow the general flow of Huber quantile loss described in \citet{dabney17}.
For two random samples $\tau, \tau' \sim U([0, 1])$,
\begin{align}
\delta_t^{\tau, \tau'} = Z_{t}^{(\lambda)}(x_t, a_t; \tau') - Z_{\theta}(x_t; \tau)
\end{align}
where $Z_t^{(\lambda)}$ is generated via SR($\lambda$) and $Z(x; \tau) = F^{-1}_Z(\tau)$ is realization of $Z(X)$ given $X=x$ and $\tau$.
Then, the full loss function of value distribution is given by
\begin{align}
\label{eq:L_IQAC}
L_{Z_\theta} = \frac{1}{N'} \sum_{i=1}^N \sum_{j=1}^{N'} \rho_{\tau_i}^\kappa \left( \delta_t^{\tau_i, \tau'_j} \right)
\end{align}
where $N$ and $N'$ are number of samples of $\tau, \tau'$, respectively, and $\rho$ is the Huber quantile loss
\begin{align}
\rho_\tau^\kappa(\delta_{ij}) &= |\tau - \mathbb{I} \{\delta_{ij} < 0\}| \frac{L_\kappa(\delta_{ij})}{\kappa}, \quad \text{with} \\
L_\kappa(\delta_{ij}) &= 
\begin{cases}
\frac{1}{2} \delta_{ij}^2, & \text{if} |\delta_{ij}| \le \kappa \\
\kappa(|\delta_{ij}| - \frac{1}{2}\kappa), & \text{otherwise}.
\end{cases}
\end{align}

\subsection{Implicit Quantile + Energy Distance (IQAC-E)}
Here, we replace the Huber quantile loss in \eqref{eq:L_IQAC} with sample-based approximation of energy distance defined in \eqref{eq:energy_distance}.
\begin{align}
L_{Z_\theta} = \frac{2}{NN'} \sum_{i=1}^N \sum_{j=1}^{N'} \left|\delta_t^{\tau_i, \tau'_j}\right| 
- \frac{1}{N^2} \sum_{i=1}^N \sum_{i'=1}^N \left|\delta_t^{\tau_i, \tau_{i'}}\right| 
- \frac{1}{{N'}^2} \sum_{j=1}^{N'} \sum_{j'=1}^{N'} \left|\delta_t^{\tau'_{j'}, \tau'_j}\right|
\end{align}

\subsection{Gaussian Mixture + Energy Distance (GMAC)}
Unlike the two previous losses, which use samples at $\tau$ generated by the implicit quantile network $Z_\theta(x_t; \tau)$, here we discuss a case in which the distribution is $k$-component Gaussian mixture parameterized with ($\mu_k, \sigma_k^2, w_k)$.

Using the expectation of a folded normal distribution, we define $\delta$ between two Gaussian distributions as
\begin{align}
\delta(\mu_i, \sigma^2_i, \mu_j, \sigma^2_j)
= \sqrt{\frac{2}{\pi}}\sqrt{\sigma_i^2 + \sigma_j^2} \exp\left({-\frac{(\mu_i - \mu_j)^2}{2(\sigma_i^2 + \sigma_j^2)}}\right)
+ (\mu_i - \mu_j) \left[ 1 - 2 \Phi \left( \frac{(\mu_i - \mu_j)}{\sqrt{2}} \right) \right].
\end{align}
Let $Z_\theta(x)$ and $Z_t^{(\lambda)}$ be Gaussian mixtures parameterized with $(\mu_{\theta i},\sigma^2_{\theta i}, w_{\theta i}), (\mu_{\lambda j}, \sigma^2_{\lambda j}, w_{\lambda j})$, respectively.
Then, the loss function for the value head is given by
\begin{align}
\begin{split}
L_{Z_\theta} = \frac{2}{NN'} &\sum_{i=1}^N \sum_{j=1}^{N'} w_{\theta i} w_{\lambda j} \delta(\mu_{\theta i}, \sigma^2_{\theta i}, \mu_{\lambda j}, \sigma^2_{\lambda j}) \\
- \frac{1}{N^2} &\sum_{i=1}^N \sum_{i'=1}^N w_{\theta i} w_{\theta i'} \delta(\mu_{\theta i}, \sigma^2_{\theta i}, \mu_{\theta i'}, \sigma^2_{\theta i'}) \\
- \frac{1}{{N'}^2} &\sum_{j=1}^{N'} \sum_{j'=1}^{N'} w_{\lambda j} w_{\lambda j'} \delta(\mu_{\lambda j}, \sigma^2_{\lambda j}, \mu_{\lambda j'}, \sigma^2_{\lambda j'}).
\end{split}
\end{align}

\newpage
\section{Pseudocode of GMAC}
\label{sec:pseudocode-gmac}
\begin{algorithm}[h]
\caption{Pseudocode of GMAC}
\label{alg:gmac}
\begin{algorithmic}
\STATE {\bfseries Input:} Initial policy parameters $\theta_0$, initial value function parameters $\phi_0$, length of trajectory $N$, number of environments $E$, clipping factor $\epsilon$, discount factor $\gamma$, weight parameter $\lambda$
\REPEAT
\FOR{$e=1$ {\bfseries to} $E$}
\STATE Collect samples of discounted sum of rewards $\{Z_1, \ldots, Z_N \}$ by running policy $\pi_k = \pi(\theta_k)$ in the environment
\STATE Compute the parameters $(\mu_i, \sigma_i, w_i)$ for each of the $\lambda$-returns $\{Z_1^{(\lambda)}, \ldots, Z_{N - 1}^{(\lambda)} \}$ by SR($\lambda$) (\autoref{alg:sr_lambda})
\STATE Compute advantage estimates $\hat{A}_t$ using GAE \citep{schulman16}, based on the current value function $V_{\phi_k}$
\ENDFOR
\STATE Gather the data from $E$ environments
\STATE Update policy using the clipped surrogate loss:
\begin{equation*}
\theta_{k+1} = \argmax_{\theta} \mathbb{E}{\left[\min \left(\frac{\pi_\theta (a_t | s_t)}{\pi_{\theta_k}(a_t | s_t)} \hat{A}_t, \ g(\epsilon, \hat{A}_t) \right)\right]}
\end{equation*}
\STATE via stochastic gradient ascent.
\STATE Update value function using the energy distance between Gaussian mixtures (\autoref{eq:energy_distance}):
\begin{equation*}
\phi_{k+1} = \argmin_{\phi} \mathbb{E} \left[ \mathcal{E}\left(V_\phi (s_t), \ Z_t^{(\lambda)} \right) \right]
\end{equation*}
\STATE via stochastic gradient descent.
\UNTIL{Final update step}
\end{algorithmic}
\end{algorithm}

The clipping function $g(\epsilon, A)$ shown in the algorithm is defined as follows:
\begin{equation*}
g(\epsilon, A) = \begin{cases}
(1 + \epsilon) A & \text{ if } A \geq 0 \\
(1 - \epsilon) A & \text{ if } A < 0
\end{cases}
\end{equation*}
Note that expectation of each loss is taken over the collection of trajectories and environments.

\newpage
\section{Implementation Details}
For producing a categorical distribution, a softmax layer was added to the output of the network. 
For producing a Gaussian mixture distribution, the mean of each Gaussian is simply the output of the network, the variance is kept positive by running the output through a softplus layer, and the weights of each Gaussian is produced through the softmax layer.
\label{sec:implementation-details}
\begin{table}[h]
\caption{Network architecture for GMAC on atari}
\label{table:network_arch}
\centering
\begin{tabular}{c|c|c|c}
\toprule
Layer Type & \multicolumn{2}{c}{Specifications} & Filter size, stride\\
\midrule
Input      & \multicolumn{2}{c}{84 x 84 x 4}  & \\
Conv1      & \multicolumn{2}{c}{20 x 20 x 32} & 8 x 8 x 32, 4 \\
Conv2      & \multicolumn{2}{c}{9 x 9 x 64}   & 4 x 4 x 64, 2 \\
Conv3      & \multicolumn{2}{c}{7 x 7 x 32}   & 3 x 3 x 32, 1 \\
FC1        & \multicolumn{2}{c}{512}    & \\
\midrule
Heads      & Policy         & Value     & \\
\hdashline \rule{0pt}{10pt}
(FC)       & action dim     & \# of modes ($=5$)   & \\
\bottomrule
\end{tabular}
\end{table}
Since our proposed method takes an architecture which only changes the value head of the original PPO network, we base our hyperparameter settings from the original paper \citep{schulman17}. We performed a hyperparameter search on a subset of variables: optimizers=\{Adam, RMSprop\}, learning rate=\{2.5e-4, 1.0e-4\}, number of epochs=\{4, 10\}, batch size=\{256, 512\}, and number of environments=\{16, 32, 64, 128\} over 3 atari tasks of Breakout, Gravitar, and Seaquest, for which there was no degrade in the performance of PPO.
\begin{table}[h]
\caption{Parameter settings for training Atari games and PyBullet tasks}
\label{table:atari_hp}
\centering
\scalebox{0.85}{\begin{tabular}{ccccc|cccc}
\toprule
Task & \multicolumn{4}{c}{Atari} & \multicolumn{4}{c}{PyBullet} \\
Parameter       & PPO & IQ & IQAC-E & GMAC & PPO & IQ & IQAC-E & GMAC \\
\midrule
Learning rate   & \multicolumn{4}{c}{2.5e-4} & \multicolumn{4}{c}{1e-4} \\
Optimizer       & \multicolumn{4}{c}{Adam} & \multicolumn{4}{c}{Adam}\\
Total frames    & \multicolumn{4}{c}{2e8} & \multicolumn{4}{c}{5e7}\\
Rollout steps   & \multicolumn{4}{c}{128} & \multicolumn{4}{c}{512}\\
Skip frame      & \multicolumn{4}{c}{4} & \multicolumn{4}{c}{1}\\
Environments    & \multicolumn{4}{c}{64} & \multicolumn{4}{c}{64}\\
Minibatch size  & \multicolumn{4}{c}{512} & \multicolumn{4}{c}{2048}\\
Epoch           & \multicolumn{4}{c}{4} & \multicolumn{4}{c}{10}\\
$\gamma$          & \multicolumn{4}{c}{0.99} & \multicolumn{4}{c}{0.99}\\
$\lambda$         & \multicolumn{4}{c}{0.95} & \multicolumn{4}{c}{0.95}\\
\hdashline \rule{0pt}{10pt}
Dirac samples   & -       & \multicolumn{2}{c}{64} & - & -       & \multicolumn{2}{c}{64} & -\\
Mixtures        & -       & -        & -         & 5 & -       & -        & -         & 5 \\
\bottomrule
\end{tabular}}
\end{table}

\newpage
\section{More Experimental Results}
\label{sec:extra_experiments}
\begin{figure}[ht]
\centering
\begin{subfigure}{.35\textwidth}
\centering
\includegraphics[scale=0.3]{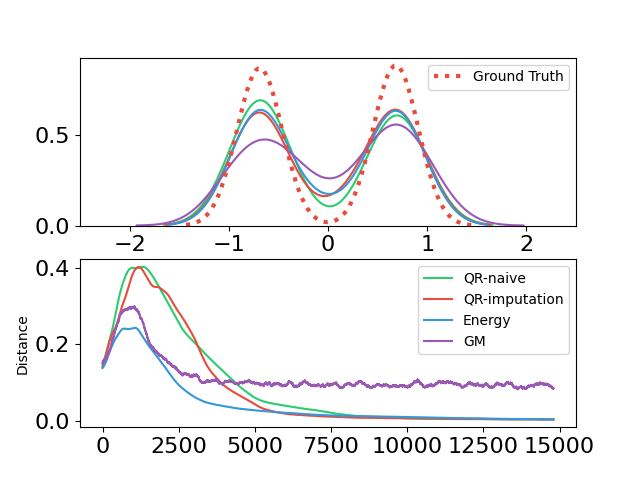}
\caption{Quantile Regression}
\label{fig:toy_qr}
\end{subfigure}
\begin{subfigure}{.35\textwidth}
\centering
\includegraphics[scale=0.3]{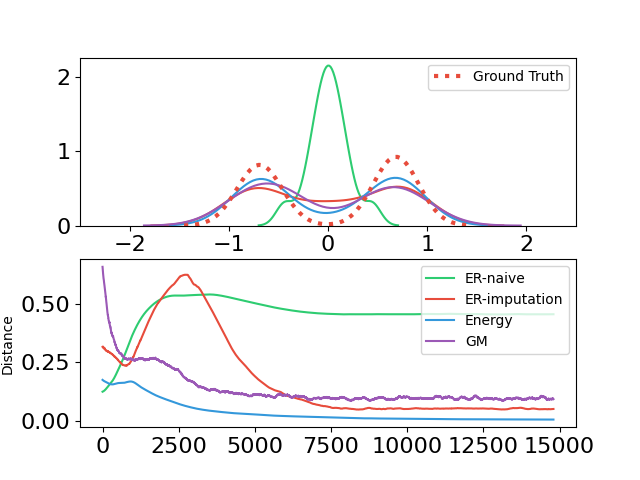}
\caption{Expectile Regression}
\label{fig:toy_er}
\end{subfigure}
\caption{In addition to \autoref{fig:modality-toy}, quantile and expectile regressions are also evaluated in the 5-state MDP against IQE and GMM with their respective loss functions.}
\end{figure}
\begin{figure}[ht]
\centering
\begin{subfigure}{.32\textwidth}
\centering
\includegraphics[scale=0.28]{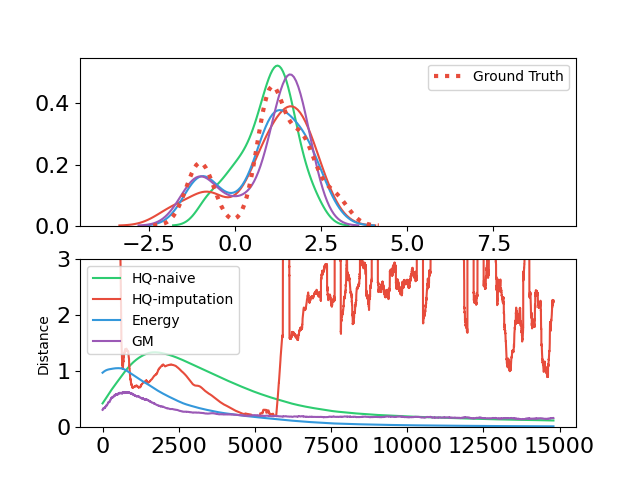}
\caption{Huber Quantile Regression}
\label{fig:toy_hq_sy}
\end{subfigure}
\begin{subfigure}{.32\textwidth}
\centering
\includegraphics[scale=0.28]{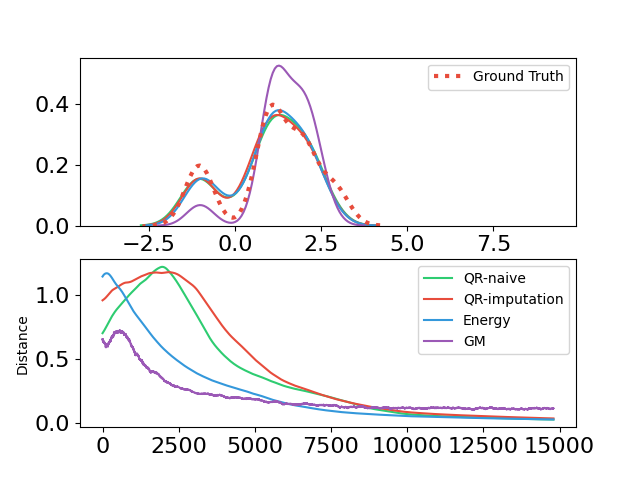}
\caption{Quantile Regression}
\label{fig:toy_qr_sy}
\end{subfigure}
\begin{subfigure}{.32\textwidth}
\centering
\includegraphics[scale=0.28]{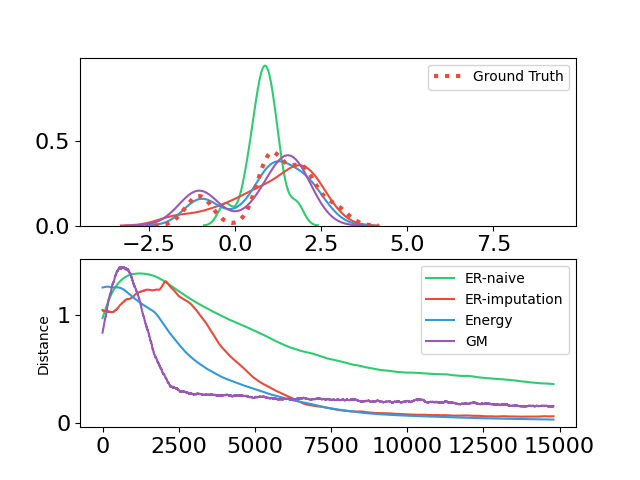}
\caption{Expectile Regression}
\label{fig:toy_er_sy}
\end{subfigure}
\caption{Evaluation of the five-state MDP under tabular setting on a asymmetric reward distribution).
Huber quantile($\kappa=1$), quantile, and expectile regressions are compared to the energy distance minimization between samples and Gaussian mixtures.}
\label{fig:tabular_toy}
\end{figure}
Here we provide more details on the five-state MDP presented in \autoref{fig:modality-toy}.
For each cases in the figure, 15 diracs are used for quantile based methods and 5 mixtures are used for GMM to balance the total number of parameters required to represent a distribution.
For the cases with the label "naive", the network outputs (quantiles, expectiles, etc.) are used to create the plot.
On the other hand, the cases with "imputation" labels apply appropriate imputation strategy to the statistics to produce samples which are then used to plot the distribution. 
Sample based energy-distance was used to calculate the distance from the true distribution for all cases. 
\begin{figure}[h]
\centering
\begin{subfigure}[b]{0.19\textwidth}
\centering
\includegraphics[scale=0.4]{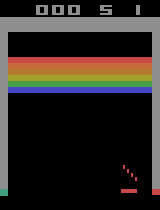}
\caption{Breakout-v4}
\end{subfigure}
\begin{subfigure}[b]{0.79\textwidth}
\centering
\includegraphics[scale=0.11]{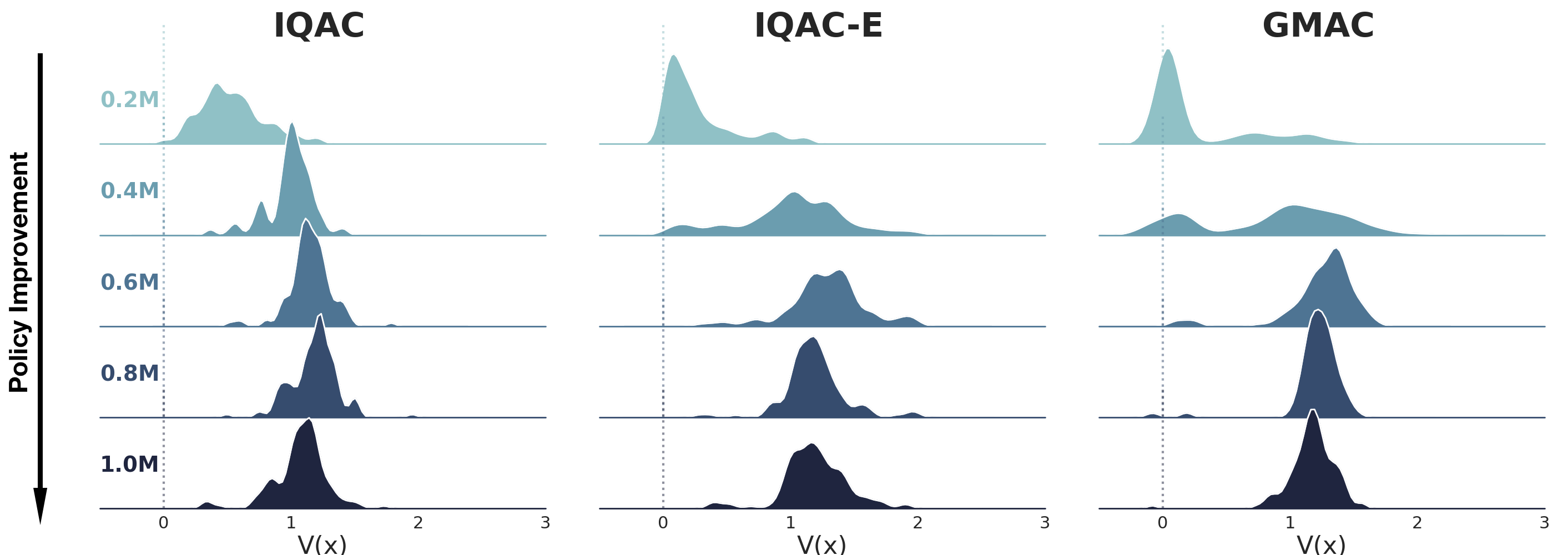}
\caption{Breakout-v4}
\end{subfigure}
\begin{subfigure}[b]{0.19\textwidth}
\centering
\includegraphics[scale=0.4]{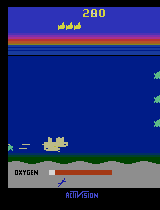}
\caption{Seaquest-v4}
\end{subfigure}
\begin{subfigure}[b]{0.79\textwidth}
\centering
\includegraphics[scale=0.11]{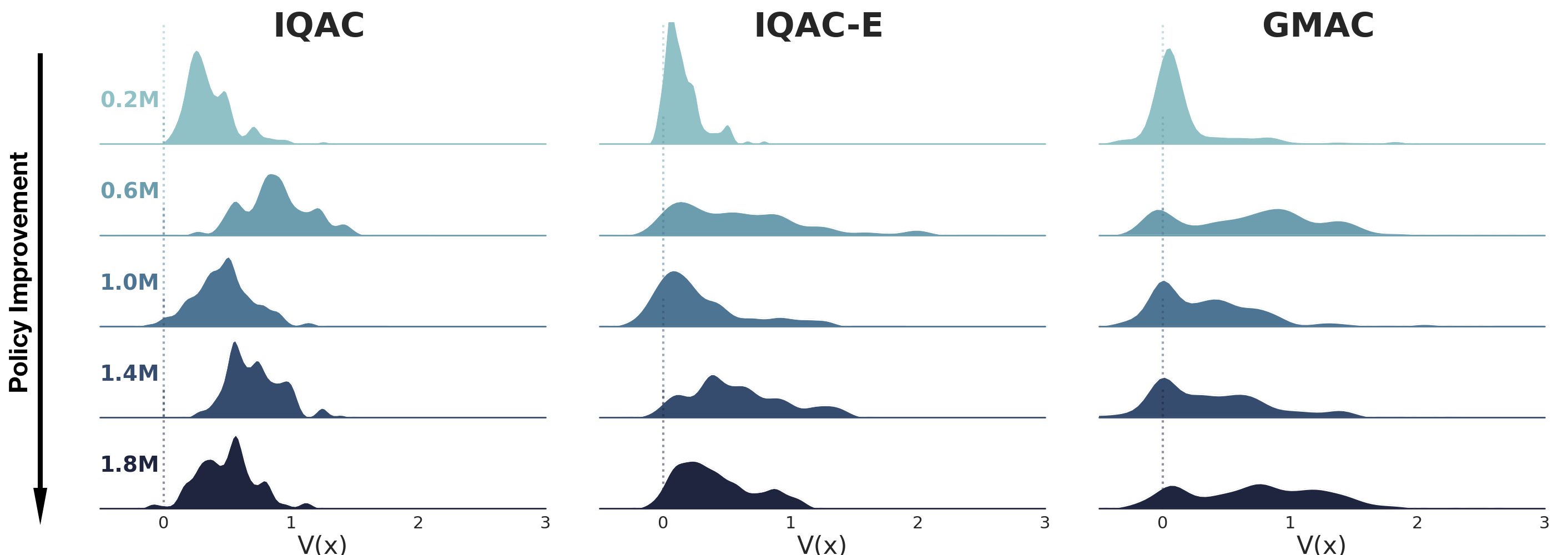}
\caption{Seaquest-v4}
\end{subfigure}
\begin{subfigure}[b]{0.19\textwidth}
\centering
\includegraphics[scale=0.4]{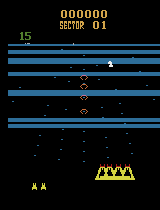}
\caption{BeamRider-v4}
\end{subfigure}
\begin{subfigure}[b]{0.79\textwidth}
\centering
\includegraphics[scale=0.11]{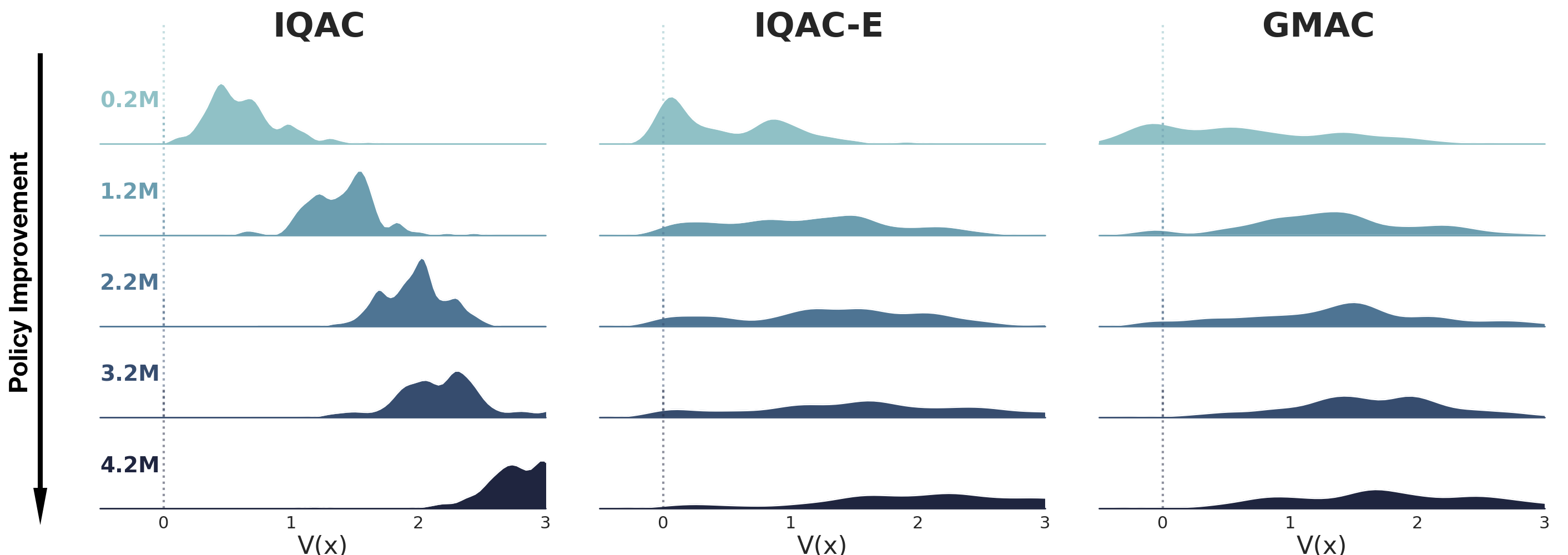}
\caption{BeamRider-v4}
\end{subfigure}
\begin{subfigure}[b]{0.19\textwidth}
\centering
\includegraphics[scale=0.4]{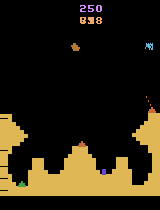}
\caption{Gravitar-v4}
\end{subfigure}
\begin{subfigure}[b]{0.79\textwidth}
\centering
\includegraphics[scale=0.11]{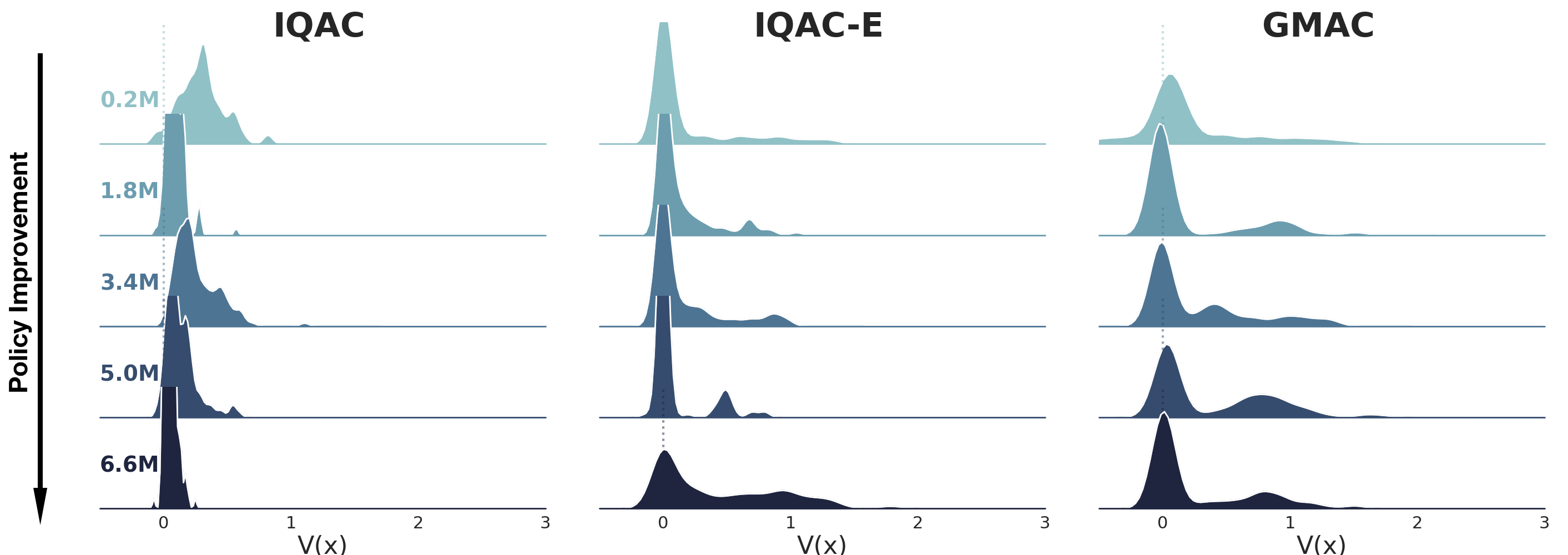}
\caption{Gravitar-v4}
\end{subfigure}
\caption{
More value distributions of different tasks.
All states are chosen such that the agent is in place of near-death or near positive score.
Thus, when the policy is not fully trained, such as in a very early stage, the value distribution should include a notion of death indicated by a mode positioned at zero. 
In all games, IQN + Huber quantile (IQAC) fails to correctly capture a mode positioned at zero while the other two methods, IQN + energy distance (IQAC-E) and GMM + energy distance (GMAC) captures the mode in the early stage of policy improvement.
Again, the visual representation is \textit{maxpool} of the 4 frame stacks in the given state.
}
\end{figure}

\begin{figure}[h]
\centering
\begin{subfigure}{.32\textwidth}
\centering
\includegraphics[scale=0.3]{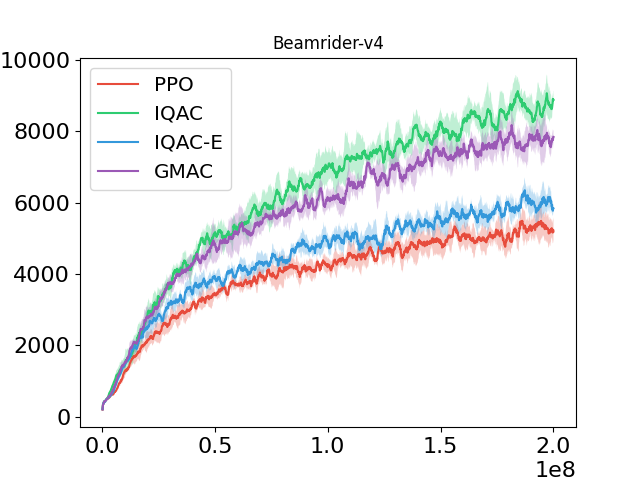}
\caption{BeamRider}
\end{subfigure}
\begin{subfigure}{.32\textwidth}
\centering
\includegraphics[scale=0.3]{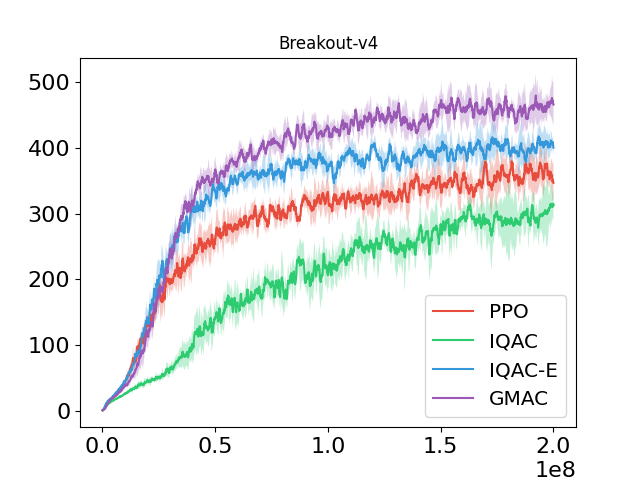}
\caption{Breakout}
\end{subfigure}
\begin{subfigure}{.32\textwidth}
\centering
\includegraphics[scale=0.3]{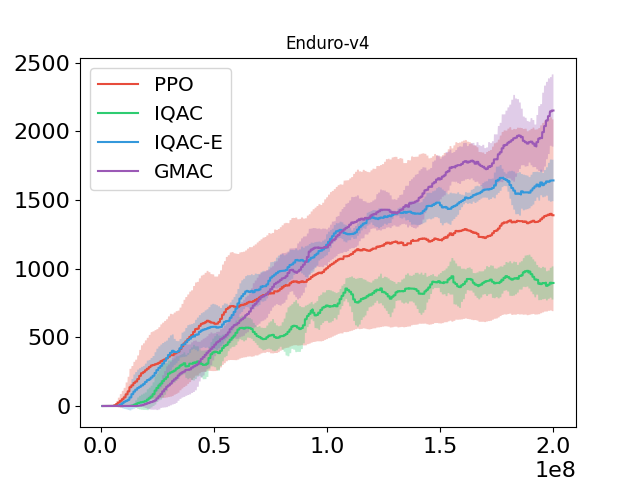}
\caption{Enduro}
\end{subfigure}
\begin{subfigure}{.32\textwidth}
\centering
\includegraphics[scale=0.3]{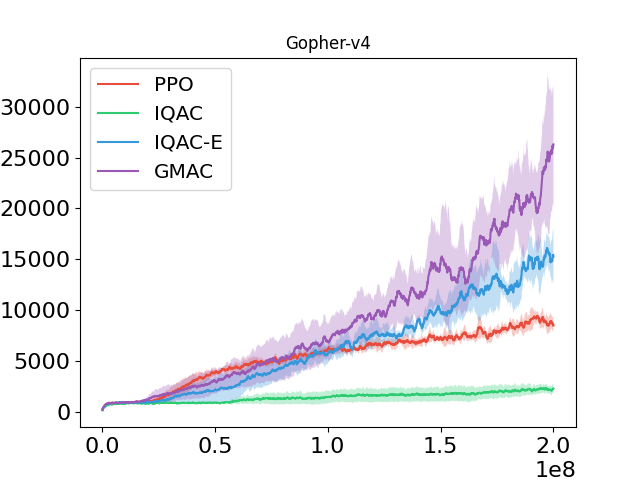}
\caption{Gopher}
\end{subfigure}
\begin{subfigure}{.32\textwidth}
\centering
\includegraphics[scale=0.3]{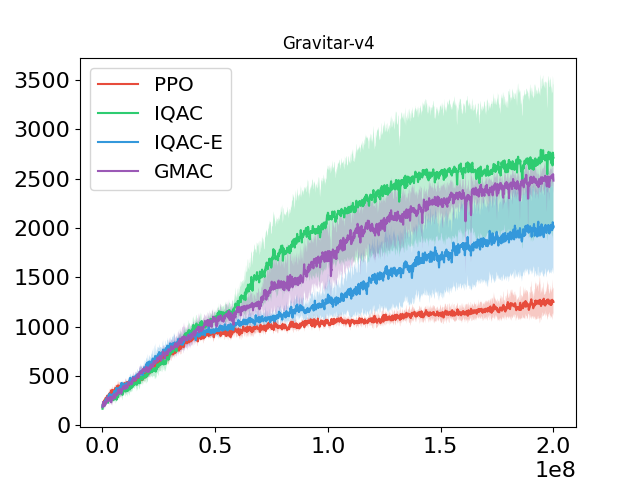}
\caption{Gravitar}
\end{subfigure}
\begin{subfigure}{.32\textwidth}
\centering
\includegraphics[scale=0.3]{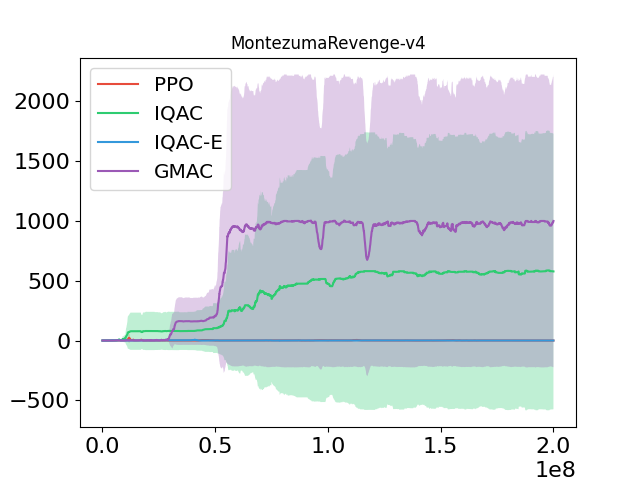}
\caption{Montezuma's Revenge}
\end{subfigure}
\begin{subfigure}{.32\textwidth}
\centering
\includegraphics[scale=0.3]{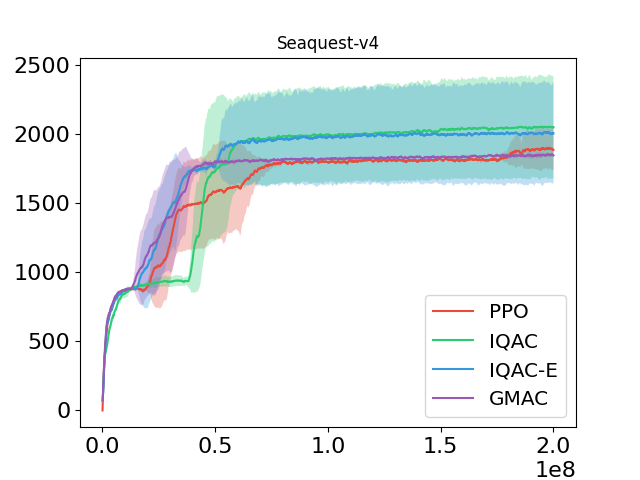}
\caption{Seaquest}
\end{subfigure}
\begin{subfigure}{.32\textwidth}
\centering
\includegraphics[scale=0.3]{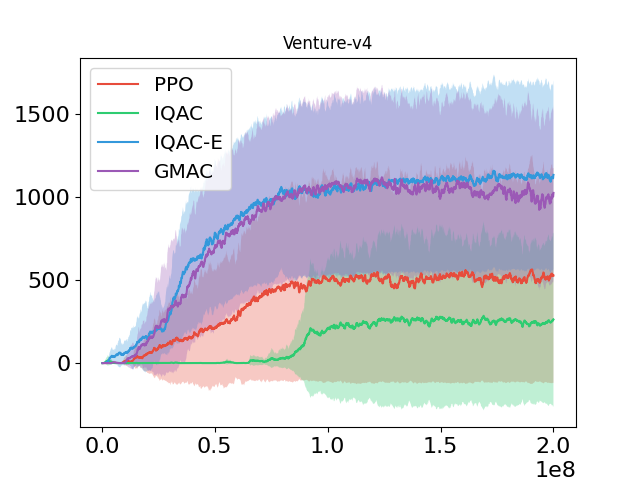}
\caption{Venture}
\end{subfigure}
\caption{
Raw learning curves over 5 random seeds for 8 selected Atari games.
The y-axis is raw score and x-axis is in frames.
The tasks are selected considering the stochasticity in games, score gap between the previous scalar and distributional method (DQN \citep{mnih15} vs. IQN \citep{dabney18}), and complexity of the game in terms of exploration.
}
\label{fig: atari-experiment-full}
\end{figure}
\begin{figure}[h]
\centering
\begin{subfigure}{.32\textwidth}
\centering
\includegraphics[scale=0.3]{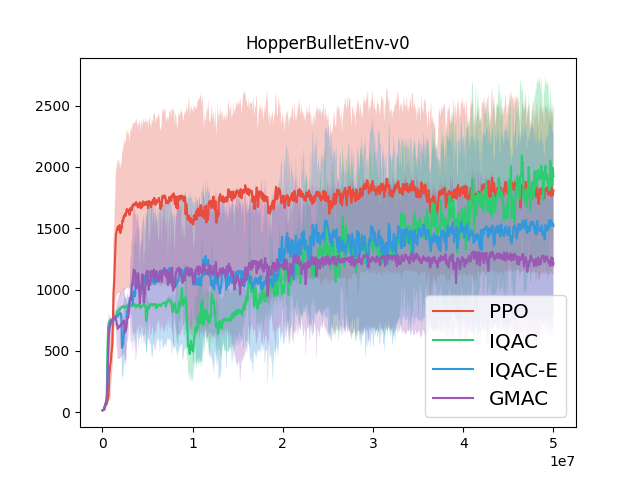}
\caption{Hopper}
\end{subfigure}
\begin{subfigure}{.32\textwidth}
\centering
\includegraphics[scale=0.3]{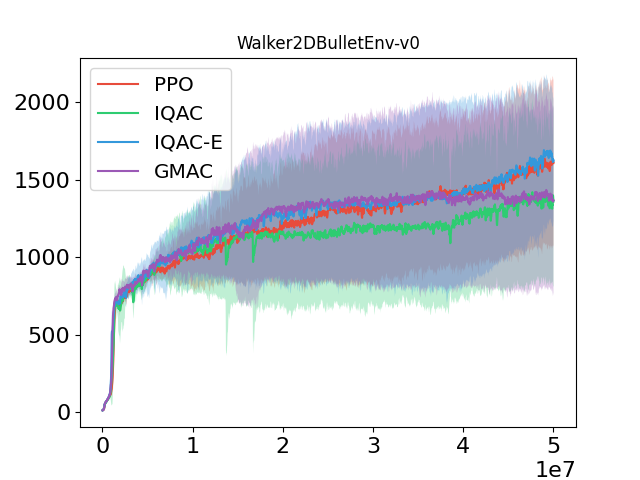}
\caption{Walker2D}
\end{subfigure}
\begin{subfigure}{.32\textwidth}
\centering
\includegraphics[scale=0.3]{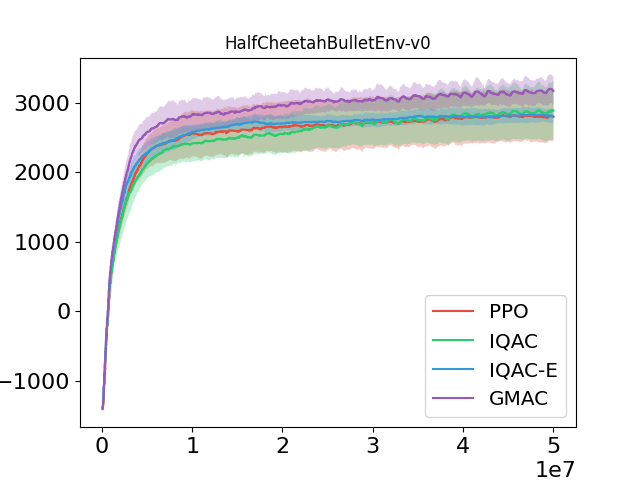}
\caption{HalfCheetah}
\end{subfigure}
\begin{subfigure}{.32\textwidth}
\centering
\includegraphics[scale=0.3]{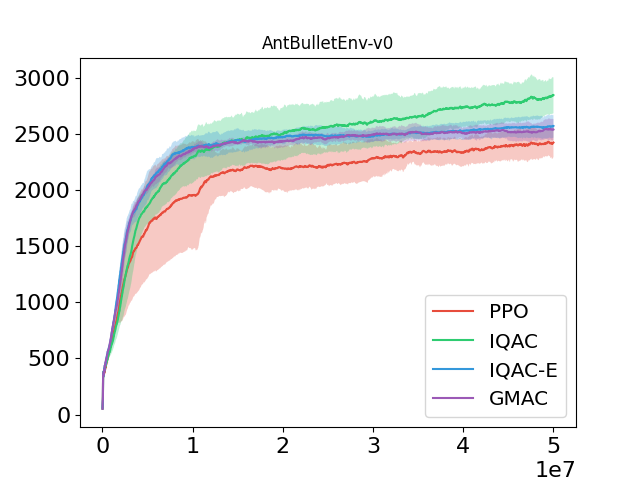}
\caption{Ant}
\end{subfigure}
\begin{subfigure}{.32\textwidth}
\centering
\includegraphics[scale=0.3]{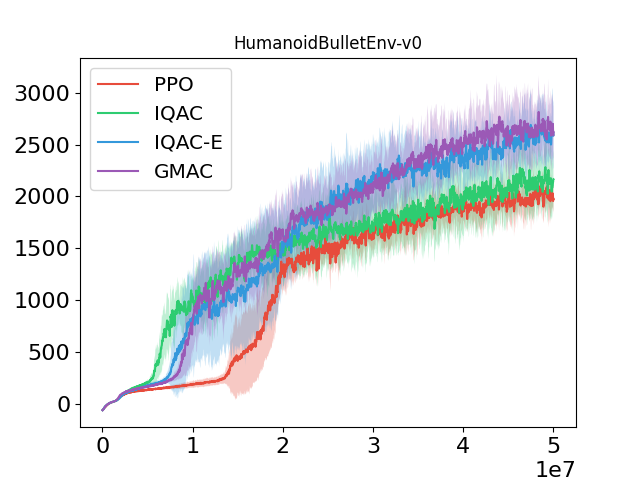}
\caption{Humanoid}
\end{subfigure}
\caption{
Raw learning curves over 5 random seeds for 5 selected PyBullet continuous control tasks.
The y-axis is in raw score and x-axis is in environment steps.
}
\label{fig: pybullet-experiment-full}
\end{figure}

\newpage
\begin{figure}[th]
\centering
\includegraphics[width=0.8\textwidth]{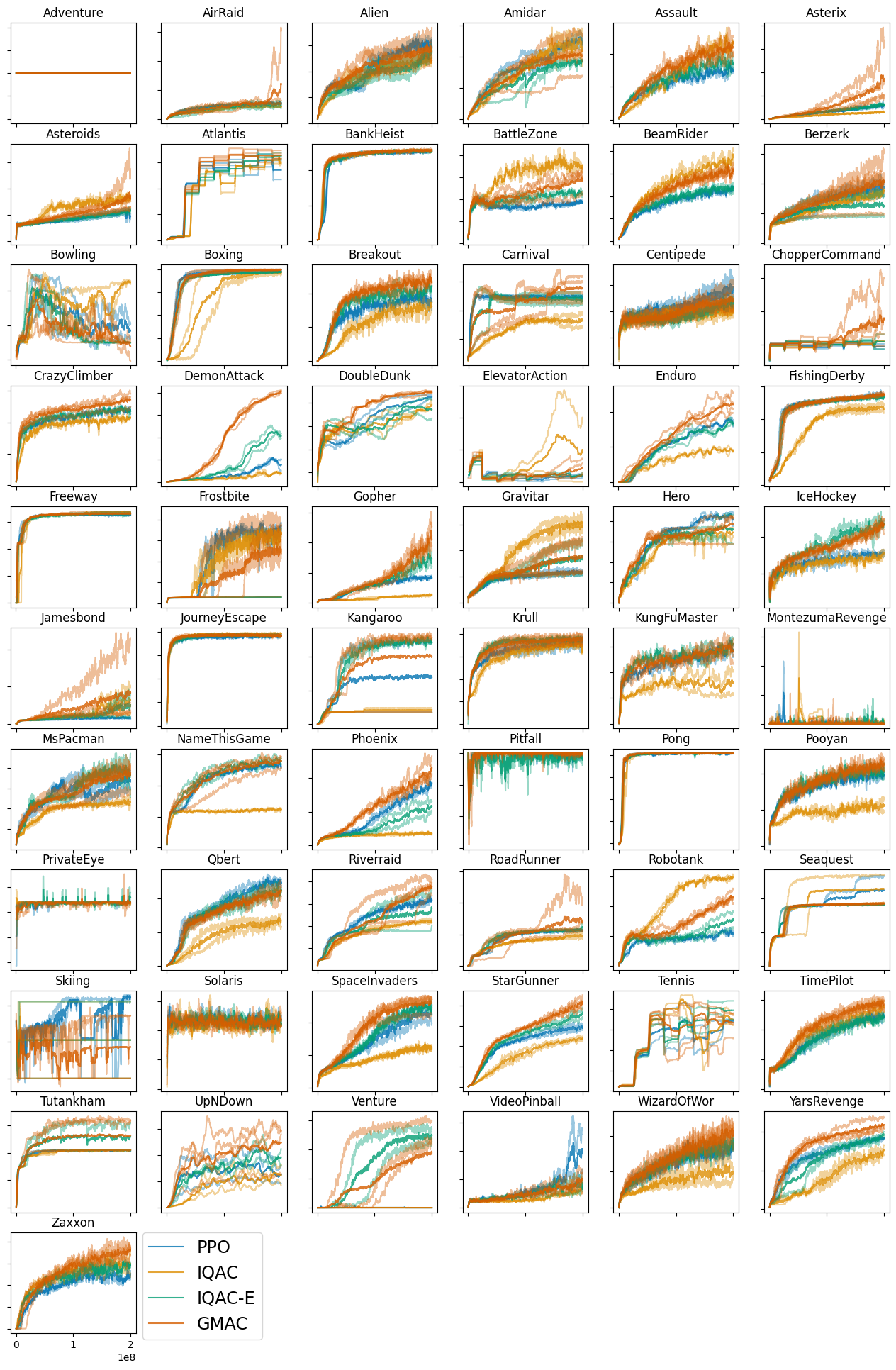}
\caption{Full learning curves of 61 atari games from ALE}
\label{fig:full_learning_curve}
\end{figure}
\newpage

\begin{table}[h]
\caption{Average score over last 100 episodes in 200M frame collected for training 61 atari games.
The algorithms are trained using same single seed and hyperparameters.
Random and Human scores are taken from Wang et al.}
\label{tab:single_seed}
\centering
\scalebox{0.82}{\begin{tabular}{c|c|c|c|c|c|c}
\textbf{GAMES}     & \textbf{RANDOM}  & \textbf{HUMAN}  & \textbf{PPO}  & \textbf{IQAC}  & \textbf{IQAC-E}  & \textbf{GMAC}  \\ 
\midrule
Adventure  & NA      & NA     & \textbf{0.00}    & \textbf{0.0}    & \textbf{0.00}   & \textbf{0.00} \\
AirRaid  & NA      & NA     & 10,205.75   & 8,304.50    & 7,589.50    & \textbf{62,328.75}  \\
Alien  & 227.80      & \textbf{7,127.70}    & 2,918.60    & 2,505.80    & 2,704.20   & 3,687.10  \\
Amidar  & 5.80      & \textbf{1,719.50}   & 1,244.12    & 1,210.40  & 932.11  & 1,363.72  \\
Assault  & 222.40      & 742.00     & 7,508.03    & \textbf{12,053.03}   & 8,589.55   & 10,281.73  \\
Asterix  & 210.00      & 8,503.30     & 13,367.00    & 6,868,00   & 15,426.00    & \textbf{22,650.00}  \\
Asteroids  & 719.10      & \textbf{47,388.70}     & 2,088.10    & 3,428.10  & 2,332.00    & 2,597.50  \\
Atlantis  & 12,850.00      & 29,028.10     & 3,073,796.00   & 2,916,292.00  & \textbf{3,373,635.00}     & 3,141,534.00  \\
BankHeist  & 14.20      & 753.10     & 1,263.80    & 1,265.80   & \textbf{1,286.60}   & 1,274.30  \\
BattleZone  & 2,360.00      & \textbf{37,187.50}     & 18,540.00    & 35,160.00 & 21,310.00  & 32,490.00  \\
BeamRider  & 363.90      & \textbf{16,926.50}     & 5,913.84    & 8,968.58  & 6,507.68  & 8,718.72  \\
Berzerk  & 123.70      & 2,630.40     & 1,748.10     & 1,682.70 & 887.50  &\textbf{3,081.20}  \\
Bowling  & 23.10      & \textbf{160.70}     & 33.54    & 65.81  & 30.00  & 19.39  \\
Boxing  & 0.10      & 12.10     & 96.79    & 97.84  & 97.10  & \textbf{99.89}  \\
Breakout  & 1.70      & 30.50     & 384.29    & 296.91  & 445.64  & \textbf{462.68}  \\
Carnival  & NA      & NA     & 5,079.20    & 2,865.40   & 4,401.00   & \textbf{6,344.20}  \\
Centipede  & 2,090.90      & \textbf{12,017.00}     & 5,205.25    & 4,085.38   & 4,864.69    & 4,303.10  \\
ChopperCommand  & 811.00      & \textbf{7,387.90}     & 872.00    & 1,096.00    & 1,314.00    & 1,795.00  \\
CrazyClimber  & 10,780.50      & 35,829.40     & 112,640.00    & 107,375.00 & 121,550.00  & \textbf{125,143.00}  \\
DemonAttack  & 152.10      & 1,971.00     & 50,590.65    & 40,369.90    & 236,839.85   & \textbf{411,118.85}  \\
DoubleDunk  & -18.60      & -16.40     & -3.26    & -6.30   & -8.28   & \textbf{-2.72}  \\
ElevatorAction  & NA      & NA     & 10,449.00    & 50.00   & 8,516.00    & \textbf{14,254.00}  \\
Enduro  & 0.00      & 860.50     & 1,588.68    & 861.65   & 1,612.17   & \textbf{2,092.65}  \\
FishingDerby  & -91.70      & -38.70     & 37.01    & 9.12   & 33.13 & \textbf{37.52}  \\
Freeway  & 0.00      & 29.60     & 32.53    & 32.96 & \textbf{33.68}    & 32.84  \\
Frostbite  & 62.50    & \textbf{4,334.70}    & 3,571.50    & 3,550.10    & 307.10  & 3,392.40  \\
Gopher  & 257.60    & 2,412.50    & 8,199.80    & 2,932.20   & 16,934.60  & \textbf{25,266.80}  \\
Gravitar  & 173.00    & \textbf{3,351.40}    & 1,151.50    & 2,798.00    & 2,178.50    & 2,401.00  \\
Hero  & 1,027.00    & 30,826.40    & 37,725.55    & 32,568.50    & \textbf{43,065.95}  & 41,509.05  \\
IceHockey  & -11.20    & 0.90    & -1.90    & -1.98  & \textbf{2.13}   & 0.34  \\
Jamesbond  & 29.00    & 302.80    & 642.50    & \textbf{4,913.50} & 961.00   & 1,512.00  \\
JourneyEscape  & NA    & NA    & -607.00    & \textbf{-339.00}   & -840.00   & -680.00  \\
Kangaroo  & 52.00    & 3,035.00    & 1,742.00    & 2,368.00  & 12,208.00 & \textbf{12,909.00}  \\
Krull  & 1,598.00    & 2,665.50    & \textbf{9,605.51}    & 8,643.09 & 9,514.03    & 9,127.63  \\
KungFuMaster  & 258.50    & 22,736.50     & 26,846.00    & 12,006.00  & 33,378.00   & \textbf{31,025.00}  \\
MontezumaRevenge  & 0.00    & \textbf{4,753.30}     & 0.00    & 3.00  & 0.00    & 0.00  \\
MsPacman  & 307.30    & \textbf{6,951.60}    & 3,674.20    & 2,450.70    & 4,699.00    & 3,884.40  \\
NameThisGame  & 2,292.30    & 8,049.00    & 13,229.10    & 6,027.80  & 13,454.00  & \textbf{14,031.30}  \\
Phoenix  & 761.40    & 7,242.60    & 37,263.70    & 6,366.20 & 26,154.00    & \textbf{42,664.00}  \\
Pitfall  & -229.40    & \textbf{6,463.70}    & 0.00    & 0.00    & -18.86    & -3.36  \\
Pong  & -20.70    & 14.60    & 20.87    & 20.68  & 20.88  & \textbf{20.97}  \\
Pooyan  & NA    & NA    & 4,018.95    & 1,819.85   & 3,674.70 & \textbf{4,178.65}  \\
PrivateEye  & 24.90    & \textbf{69,571.30}    & 100.00    & 71.51   & 196.30  & 100.00  \\
Qbert  & 163.90    & 13,455.00     & \textbf{25,519.25}    & 11,728.25    & 21,599.50   & 23,176.25  \\
Riverraid  & 1,338.50    & 17,118.00    & 15,983.00    & 10,840.80 & 18,073.40   & \textbf{19,761.30}  \\
RoadRunner  & 11.50    & 7,845.00    & 56,321.00    & 44,685.00  & 56,121.00  & \textbf{68,272.00}  \\
Robotank  & 2.20    & 11.90    & 23.45    & \textbf{60.79}    & 36.69    & 45.82  \\
Seaquest  & 68.40      & \textbf{42,054.70}    & 1,832.00    & 2,704.40    & 1,814.60    & 1,838.40  \\
Skiing  & -17,098.10      & \textbf{-4,336.90}     & -7,958.81    & -8,987.12   & -29,971.02  & -29,975.52  \\
Solaris  & 1,236.30    & \textbf{12,326.70}    & 2,452.80    & 2,342.60  & 2,204.80    & 2,579.20  \\
SpaceInvaders  & 148.00     & 1,668.70     & \textbf{2,544.10}    & 1,177.65    & 2,410.90    & 2,228.30  \\
StarGunner  & 664.00      & 10,250.00     & 74,848.00    & 57,053.00    & 97,450.00   & \textbf{104,188.00}  \\
Tennis  & -23.80    & -8.30    & -8.16    & -6.17    & -7.54    & \textbf{-5.90}  \\
TimePilot  & 3,568.00      & 5,229.20     & 12,157.00    & \textbf{14,746.00}    & 11,704.00   & 13,227.00  \\
Tutankham  & 11,40     & 167.60     & 206.32    & \textbf{210.66}    & 208.72    & 209.82 \\
UpNDown  & 533.40    & 11,693.20     & 158,629.50    & 84,962.70    & \textbf{161,328.40} & 129,243.70  \\
Venture  & 0.00    & 1,187.50    & 0.00   & 0.00    & \textbf{1,339.00} & 1,181.00  \\
VideoPinball  & 16,256.90      & 17,667.90     & \textbf{279,504.81}    & 55,113.30 & 59,988.90  & 55,272.82  \\
WizardOfWar  & 563.50      & 4,756.50     & 8,749.00    & 5,688.00  & 9,165.00  & \textbf{11,388.00}  \\
YarsRevenge  & 3,092.90      & 54,576.90     & 92,709.94    & 83,136.68 & 100,082.55  & \textbf{103,895.05}  \\
Zaxxon  & 32.50      & 9,173.00     & 13,336.00    & 11,886.00  & 14,882.00  & \textbf{18,436.00}  \\
\end{tabular}}
\end{table}

\end{appendices}

\end{document}